\documentclass[10pt,twocolumn,letterpaper]{article}

\usepackage{iccv}
\usepackage{times}
\usepackage{epsfig}
\usepackage{graphicx}
\usepackage{amsmath}
\usepackage{amssymb}
\usepackage{bm}
\usepackage{caption}
\usepackage{color}
\usepackage{xcolor}
\usepackage{enumitem}
\usepackage{multirow}
\usepackage{makecell}

\DeclareMathOperator*{\argmax}{arg\,max}

\usepackage[pagebackref=true,breaklinks=true,colorlinks,bookmarks=false]{hyperref}
\hypersetup{linkcolor=[rgb]{0.8,0.1,0.1}}
\hypersetup{citecolor=[rgb]{0.4,0.15,0.95}}

\iccvfinalcopy 

\ificcvfinal\fi

\begin{document}

\title{\vspace{-6mm}Self-Supervised 3D Face Reconstruction via Conditional Estimation}

\author{Yandong Wen\textsuperscript{1}~~~~~Weiyang Liu\textsuperscript{2,3}~~~~~Bhiksha Raj\textsuperscript{1}~~~~~Rita Singh\textsuperscript{1}\\
\textsuperscript{1}Carnegie Mellon University~~~~\textsuperscript{2}University of Cambridge~~~~\textsuperscript{3}MPI for Intelligent Systems, Tübingen\\
}

\maketitle
\ificcvfinal\thispagestyle{empty}\fi

\begin{abstract}
   We present a conditional estimation (CEST) framework to learn 3D facial parameters from 2D single-view images by self-supervised training from videos. CEST is based on the process of analysis by synthesis, where the 3D facial parameters (shape, reflectance, viewpoint, and illumination) are estimated from the face image, and then recombined to reconstruct the 2D face image. In order to learn semantically meaningful 3D facial parameters without explicit access to their labels, CEST couples the estimation of different 3D facial parameters by taking their statistical dependency into account. Specifically, the estimation of any 3D facial parameter is not only conditioned on the given image, but also on the facial parameters that have already been derived. Moreover, the reflectance symmetry and consistency among the video frames are adopted to improve the disentanglement of facial parameters. Together with a novel strategy for incorporating the reflectance symmetry and consistency, CEST can be efficiently trained with in-the-wild video clips. Both qualitative and quantitative experiments demonstrate the effectiveness of CEST.
\end{abstract}

\vspace{-3.3mm}
\section{Introduction}
\vspace{-0.2mm}
Reconstructing 3D faces from single-view 2D images has been a longstanding problem in computer vision. The common approach represents the 3D face as a combination of its {\em shape}, as represented by the 3D coordinates of a number of points on its surface called vertices, and its {\em texture}, as represented by the reflectances of red, green and blue at these vertices \cite{blanz1999morphable}. The problem then becomes learning a regression model between the 2D images, and vertices and their reflectances.

\begin{figure}[t]
  \centering
  \setlength{\abovecaptionskip}{3pt}
 \setlength{\belowcaptionskip}{-10pt}
  \renewcommand{\captionlabelfont}{\footnotesize}
 \includegraphics[width=3.25in]{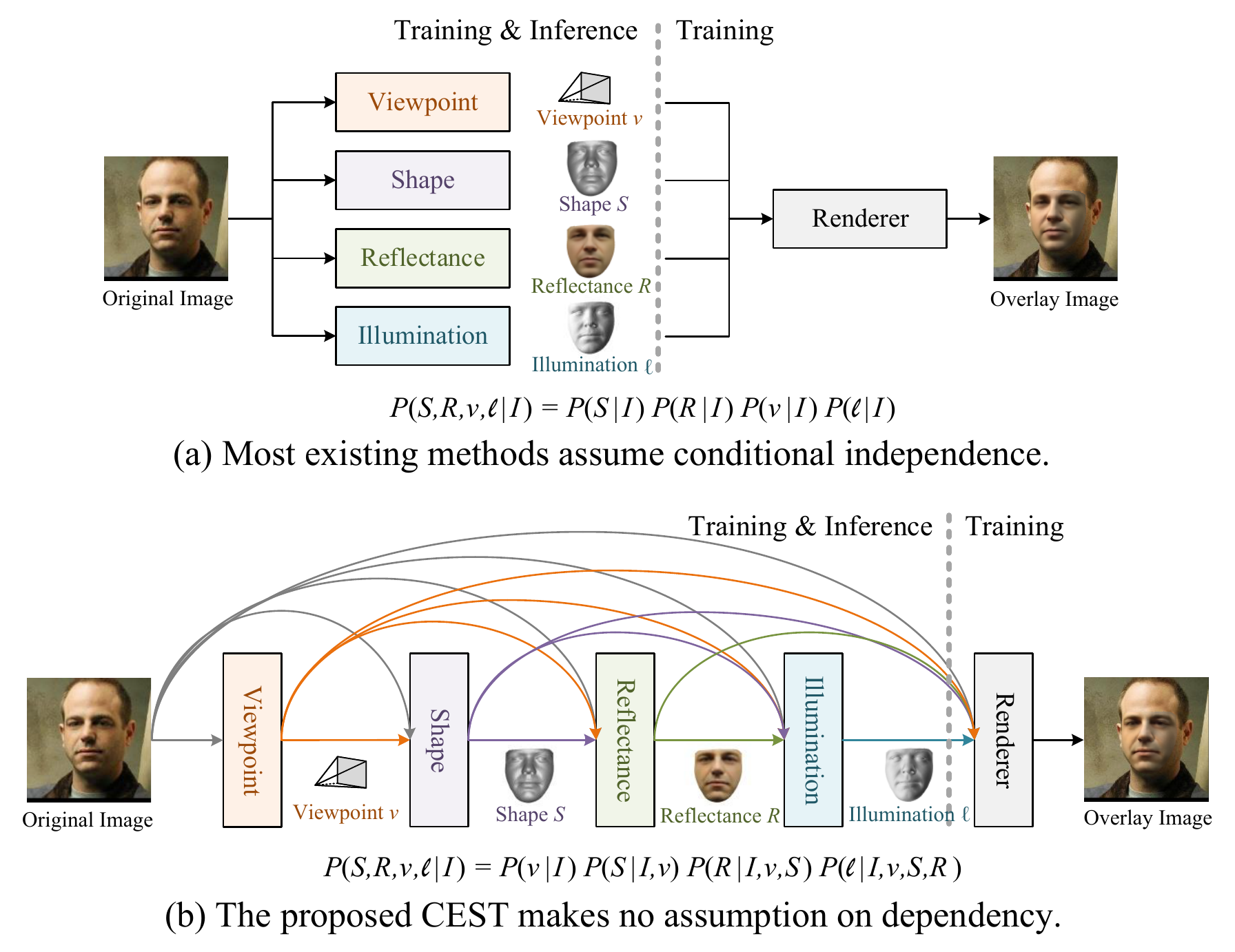}
  \caption{\footnotesize Conventional 3D face reconstruction and our CEST framework. The dotted lines separate the modules used for inference of the 3D parameters from those used for training with self-supervision.}\label{fig:overview}
\end{figure}

The regression itself may be learned using training data where both, the 2D images and the corresponding 3D parameters are available. However, these data are scarce, and even the ones that are available generally only have shape information \cite{cao2013facewarehouse,zhang2013high, yin20063d}; the ones that do have other parameters are usually captured in a controlled environment \cite{lattas2020avatarme} or are synthetic \cite{sengupta2018sfsnet}, which is not representative of real-world images. Consequently, there is great interest in self-supervised learning methods, which learn the regression model from natural in-the-wild 2D images or videos, without explicit access to 3D training data \cite{tewari2017mofa,tran2018nonlinear}.

The problem is complicated by the fact that the actual image formation depends not only on the shape and texture of the face, but also the illumination (the intensity and direction of the incident light), and other factors such as the viewpoint (incorporating the orientation of the face and the position of the camera), etc. Thus, the learned regression model must also account for these factors. To this end, the general approach is one where shape, reflectance, illumination and viewpoint parameters are all extracted from the 2D image. The regression model that extracts these facial parameters are learned through {\em self-supervision}: the extracted facial parameters are recombined to render the original 2D image, and the model parameters are learned to minimize the reconstruction error. 

The solution, however, remains ambiguous because a 2D image may be obtained from different combinations of shape, texture, illumination and viewpoint. To ensure that the self-supervision provides meaningful disentanglement, the manner in which the facial parameters are recombined to reconstruct the 2D image are based on the actual physics of image formation~\cite{tewari2017mofa, tran2018nonlinear, sengupta2018sfsnet}. To further reduce potential ambiguities, regularizations are necessary. Reflectance {\em symmetry} has been proposed as a regularizer \cite{tran2019learning, tewari2018self, wu2020unsupervised}, wherein the reflectance of a face image and its mirror reflection are assumed to be identical. Smoothness has also been employed to regularize the shape and reflectance~\cite{tran2018nonlinear, tewari2018self}.  Additional regularization may be obtained by considering correspondences between multiple images of the same face \cite{kemelmacher2011face,tewari2019fml}, particularly when they are obtained under near identical conditions such as the sequence of images from a video. The approach in \cite{tewari2019fml} has considered reflectance {\em consistency}, where reflectances of all image frames in a video clip are assumed to be similar.

\begin{figure*}[t]
  \centering
  \setlength{\abovecaptionskip}{0pt}
 \setlength{\belowcaptionskip}{-7pt}
  \renewcommand{\captionlabelfont}{\footnotesize}
 \includegraphics[width=6.6in]{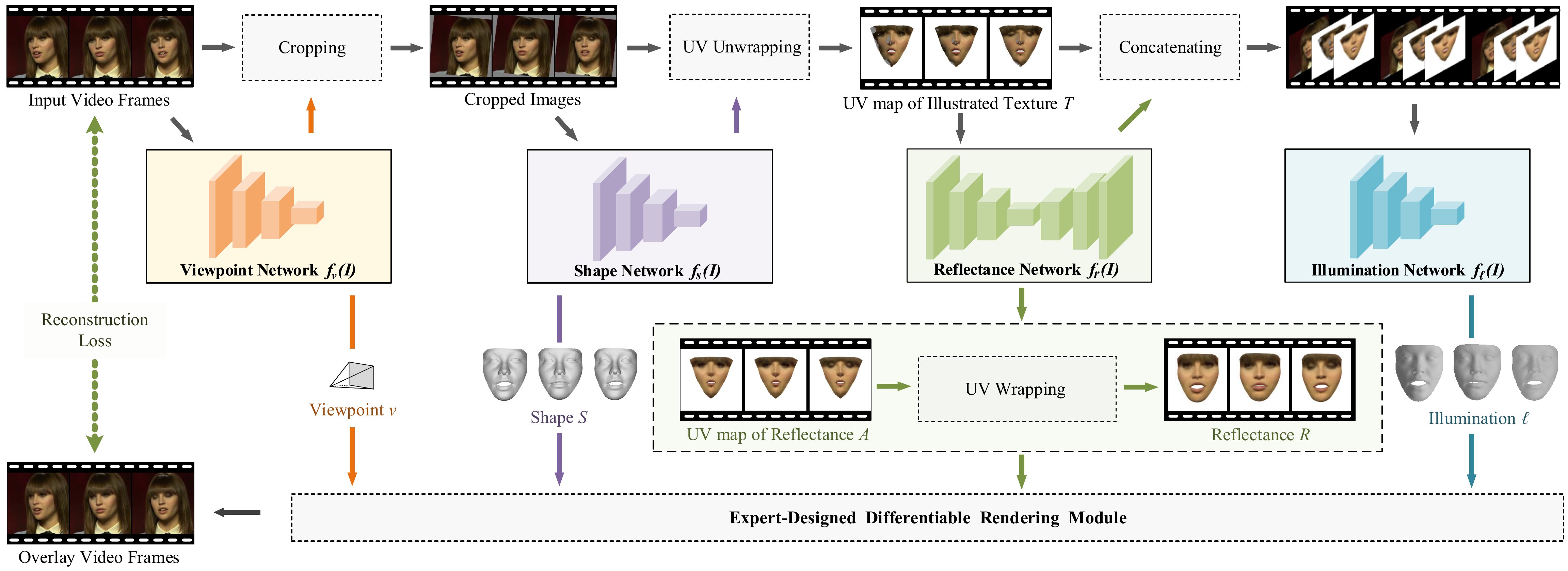}
  \caption{\footnotesize The overall training pipeline of the proposed CEST framework.}\label{fig:cest_framework}
\end{figure*}

In all of these prior works, the target parameters, namely the shape, reflectance, illumination and viewpoint parameters are all {\em individually} estimated, without considering their direct influences on one another, although they are jointly optimized. In effect, at inference time they assume that the estimate of, {\em e.g.} the reflectance, is conditionally independent of the estimated shape or viewpoint, given the original 2D image. The coupling among the four is only considered during (self-supervised) training, where they must all combine to faithfully recreate the input 2D image \cite{garrido2016reconstruction,genova2018unsupervised,richardson2017learning,tran2019learning,tewari2019fml}. This is illustrated in Fig. \ref{fig:overview}(a). 

In reality, 2D images are reduced-dimensional projections, and thus imperfect representations of the full three-dimensional structure of the face, and the aspects of reflectance and illumination imprinted in them are not independent of the underlying shape of the object or the viewpoint they were captured from. Therefore, the captured 2D image represents a joint interaction among viewpoint, shape, reflectance and illumination. Consequently, the statistical estimates of any of these four factors may not, in fact, be truly conditionally independent of one another given only the 2D image (although, given the entire 3D model they might have been). Thus, modelling all of these variables as being conditionally independent effectively represents a lost opportunity since, by predicting them individually, the constraints they impose on one another are ignored. Optimization-based approaches~\cite{kemelmacher20103d, kemelmacher2011face,shi2014automatic} do attempt to capture the dependence by iteratively estimating shape and reflectance from one another. However, these methods require correspondence information of the image sequence in a video and suffer from costly inference.

In this paper, we propose a novel learning-based framework based on conditional estimation (CEST). CEST explicitly considers the statistical dependency of the various 3D facial parameters (shape, viewpoint, reflectance and illumination) upon one another, when derived from single 2D image. The specific form of the dependencies adopted in this paper is shown in Fig.~\ref{fig:overview}(b). We note that the CEST framework is very general and allows us to consider any other dependency structures. Our paper serves as one of the many potential choices that work well in practice. To this end, we present a specific, and intuitive, solution in CEST, where the viewpoint, facial shape, facial reflectance, and illumination are predicted {\em sequentially} and {\em conditionally}. In this context, the prediction of facial shape is conditioned on the input image and the derived viewpoint; the prediction of facial reflectance is conditioned on the input image, derived viewpoint and facial shape; and so forth.

As before, learning remains self-supervised, through comparison of re-rendered 2D images obtained with the estimated 3D face parameters to the original images. As additional regularizers, we also employ reflectance symmetry constraints \cite{tran2019learning, tewari2018self, wu2020unsupervised}, and  reflectance consistency constraints (across frames in a short video clip) \cite{tewari2019fml}. These are included in the form of cross-frame reconstruction error terms, the number of which increases quadratically with the number of video frames considered together for self-supervision. To address the dramatically increased number of reconstruction terms, we propose a stochastic optimization strategy to improve training efficiency.

We present ablation studies and comparisons to state-of-the-art methods \cite{tewari2017mofa,tran2019learning,tewari2019fml} to evaluate CEST. We show that CEST produces better reflectance and structured illumination, leading to more realistic rendered faces with fine facial details, compared to all other tested methods. It also achieves better shape estimation accuracy on AFLW2000-3D \cite{zhu2016face} and MICC \cite{bagdanov2011florence} datasets than current state-of-the-art self-supervised and {\em fully supervised} approaches. Overall, our contributions can be summarized as follows:

\begin{itemize}[leftmargin=*,nosep,nolistsep]
\item We propose CEST, a conditional estimation framework for 3D face reconstruction that explicitly considers the statistical dependencies among 3D face parameters.

\item We propose a specific design for the decomposition of conditional estimation, where the viewpoint, shape, reflectance, and the illumination are derived sequentially.

\item We propose a stochastic optimization strategy to efficiently incorporate reflectance symmetry and consistency constraints into CEST. As the number of video frames increase, the computational complexity of CEST is increased linearly, rather than quadratically.

\end{itemize}

\vspace{-1mm}
\section{Related Work}
\vspace{-1mm}
\textbf{Monocular 3D face reconstruction by self-supervised learning.} Many research studies published recently aim to learn 3D facial parameters from a single image in a self-supervised manner. In \cite{richardson2017learning}, the authors propose a coarse-to-fine framework to improve the details in reconstructed 3D faces. Ayush \emph{et al.} \cite{tewari2017mofa} present a model-based deep convolutional face autoencoder (MoFA) to fit a 3DMM to shape, reflectance, and illuminance. InverseFaceNet \cite{kim2018inversefacenet} trains a direct regression model on a synthetic training corpus that is generated by self-supervised bootstrapping. SfSNet \cite{sengupta2018sfsnet} combines labeled synthetic and unlabeled real-world images in learning, and produces accurate depth map, and reflectance and shade disentanglement. To better characterize facial details, 3DMM is generalized to a nonlinear model in \cite{tran2018nonlinear, tran2019learning}. \cite{zhou2019dense} uses mesh convolutions for 3D faces, leading to a light-weight model with competitive performance. \cite{shang2020self} incorporates the multi-view consistency from geometry, pixel, and depth as constraints.

However, these approaches generally do not consider correspondences across frames in a video. FML \cite{tewari2019fml} is the first self-supervised framework that incorporates  video clues in training. The shape and reflectance for each video frame are approximated by averaging the shapes and reflectances in a video clip. However,  models trained on the averaged representations may not work well for a single image if the number of multi-frame images is large, due to the large gap between averaged and isolated images.  On the contrary, CEST uses representations from single images. More importantly, it uses conditional estimation for predicting the facial parameters, and does not assume conditional independence between them, an often unrealistic assumption employed in the previously mentioned approaches.

\textbf{Optimization-based 3D face reconstruction.} \cite{kemelmacher2011face} proposes to fit a template model to
photo-collections by updating the viewpoint, geometry, lighting, and texture iteratively. \cite{shi2014automatic} fits a face model to detected 3D landmarks, and refines the texture and geometry details. \cite{garrido2016reconstruction} learns facial subspaces for identity and expression variations with a parametric shape prior. \cite{garg2013dense} considers 3D face reconstruction as a global variational energy minimization problem, and estimates dense low-rank 3D shapes for video frames.

While these approaches can be considered conditional estimation,  they focus on deriving 3D facial parameters from video, and are not relevant to the problem of deriving them from single-frame images, the problem addressed in our work. For CEST, video clips are viewed as consistent collections of images used to better learn the model. 

\vspace{-1.4mm}
\section{The CEST Framework}
\vspace{-0.6mm}
In this work, we adopt a common practice from 3D Morphable Model (3DMM) \cite{blanz1999morphable}, which represents a 3D face as a combination of shape and reflectance. The shape comprises a collection of vertices $\bm{S} = [\bm{S}(1); \bm{S}(2); ...; \bm{S}(K)] \in \mathbb{R}^{K \times 3}$, where $K$ is the number of vertices and $\bm{S}(i) = [\bm{S}(i, 1), \bm{S}(i, 2), \bm{S}(i, 3)]$ denotes the $xyz$ coordinates in the Cartesian coordinate system. The typology for $\bm{S}$ is consistent for different faces. The reflectance comprises a collection of pixel values $\bm{R} = [\bm{R}(1); \bm{R}(2); ...; \bm{R}(K)] \in \mathbb{R}^{K \times 3}$. Each row $\bm{R}(i) = [\bm{R}(i, 1), \bm{R}(i, 2), \bm{R}(i, 3)]$ comprises the pixel values (\ie, RGB) at position $\bm{S}(i)$. 
\vspace{-0.8mm}
\subsection{Framework Overview}
\vspace{-0.8mm}
\def \R {\bm{R}}
\def \S {\bm{S}}
\def \l {\bm{\ell}}
\def \v {\bm{v}}
\def \I {\bm{I}}

The problem of 3D face reconstruction from a 2D image is that of obtaining estimates of the shape $\S$, reflectance $\R$, viewpoint $\v$ and illumination $\l$, given an input image $\I$.  Statistically, we aim to estimate the most likely values for these variables, given the input image:
\begin{equation}
\footnotesize
\begin{aligned}
\hat{\S},\hat{\R},\hat{\v},\hat{\l} = \arg\max_{\S,\R,\v,\l} P(\S,\R,\v,\l | \I)
\end{aligned}
\end{equation}
The challenges of this estimation are twofold: first $P(\S,\R,\v,\l|\I)$ must be modelled, and second, $\argmax_{\S,\R,\v,\l} P(\S,\R,\v,\l | \I)$ must be computed.

Modelling $P(\S,\R,\v,\l | \I)$ directly is a challenging problem, and the problem must be factored down. Prior approaches \cite{tran2018nonlinear,tewari2017mofa,zhou2019dense} have decomposed this problem by assuming that shape, reflectance, viewpoint and illumination are all conditionally independent, given the image, i.e. $P(\S,\R,\v,\l|\I) = P(\S|\I)P(\R|\I)P(\v|\I)P(\l|\I)$.  This leads to simplified estimates where each of the variables can be independently estimated, i.e. $\hat{\S} = \argmax_{\S} P(\S | \I)$,  $\hat{\R} = \argmax_{\R} P(\R | \I)$, etc.  As we have discussed earlier, the conditional independence assumption is questionable, since the conditioning variable, $\I$, is a lower-dimensional projection of the 3D face that entangles the four variables.

In CEST we explicitly model the conditional dependence, as shown in Fig.~\ref{fig:overview}(b).  Specifically we decompose the joint probability as
\begin{equation}
\footnotesize
\begin{aligned}
& P(\S,\R,\v,\l | I) \\
\ \  = & \ P(\v|\I) P(\S| \I, \v) P(\R | \I, \v, \S) P(\l | \I, \v, \S, \R)
\end{aligned}
\end{equation}
Coupling the variables in this manner results in a complication: even factored as above, maximizing the joint probability with respect to $\S$, $\R$, $\v$, and $\l$ must be jointly performed, since the variables are coupled. We approximate it instead with the following sequential estimate, based on the sequential decomposition above:
\begin{equation}
\footnotesize
\begin{aligned}
 \hat{\v} &= \argmax_{\v} P(\v|\I) & \hat{\S} & = \argmax_{\S} P(\S | \I, \hat{\v}) \\
\hat{\R} &= \argmax_{\R} P(\R | \I, \hat{\bm{v}}, \hat{\S}) &  \hat{\l} &=  \argmax_{\l} P(\l | \I, \hat{\v}, \hat{\S}, \hat{\R}) 
\end{aligned}
\end{equation}

The second challenge is that of actually computing the $\argmax$ operations in Equation 3. Rather than attempting to model the probability distributions explicitly and maximizing them, we will, instead,  model the estimators in Equation 3 as parametric functions:
\begin{equation}
\footnotesize
\begin{aligned}
 \hat{\v} &= f_v(\I; \theta_v) &  \hat{\S}  &= f_s(\I, \hat{v}; \theta_s)  \\
 \hat{\R} &= f_r(\I, \hat{v}, \hat{\S}; \theta_r) &   \hat{\l} &= f_{\ell}(\I, \hat{\v}, \hat{\S}, \hat{\R}; \theta_{\l}) 
\end{aligned}
\end{equation}
The problem of {\em learning}  to estimate the 3D facial parameters thus effectively reduces to that of estimating the parameters $\theta_v$, $\theta_s$, $\theta_r$ and $\theta_l$.

Using the common approach, we formulate the learning process for these parameters through an autoencoder. $f_v()$, $f_s()$, $f_r()$ and $f_{\ell}()$ are, together, viewed as the learnable encoder in the autoencoder, which estimate $\v$, $\S$, $\R$ and $\l$ respectively.  The decoder is a deterministic differentiable {\em renderer} $\mathcal{R}()$ with no learnable parameters, which reconstructs the original input $\I$ from the values derived by the encoder as $\hat{\bm{I}} = \mathcal{R}(\bm{S}, \bm{R}, \bm{v}, \bm{\ell})$. The parameters of the encoder are learned to minimize the error between $\hat{\bm{I}}$ and $\I$.

\vspace{-0.8mm}
\subsection{Facial Parameters Inference}
\vspace{-0.8mm}
\textbf{Viewpoint}. We first predict the viewpoint parameters from the given image, using a function $f_v(\bm{I}; \bm{\theta}_v):\bm{I} \rightarrow \bm{v} \in \mathbb{R}^7$. Here $\bm{v}$ is used to parameterize the weak perspective transformation \cite{szeliski2010computer}, including 3D spatial rotation (SO(3)), the translation ($xyz$ coordinates), and the scaling factor.

\textbf{Shape}. The prediction of shape is conditioned on the given image $\bm{I}$ and the predicted $\bm{v}$. Since the same face captured with different viewpoints should be correspond to the same facial shape, it is beneficial to exclude as much viewpoint information from the image $\bm{I}$ as possible before the shape prediction. With the predicted $\bm{v}$, we can align the image to its canonical view in 2D plane, as shown in Fig. \ref{fig:cest_framework} and Appendix \ref{cropping}. The cropped image is denoted by $\bm{I}\circ \bm{v}$. A function $f_s(\bm{I}\circ \bm{v}; \bm{\theta}_s):\bm{I}\circ \bm{v} \rightarrow \bm{\alpha}\in \mathbb{R}^{228\times 1}$ with learnable parameter $\bm{\theta}_s$ is constructed to predict the shape coefficients $\bm{\alpha}$. The shape coefficients $\bm{\alpha}$ are defined by a statistical model of 3D facial shape:
\begin{equation}
\footnotesize
\vec{\bm{S}} = \bar{\bm{S}} + \bm{U}\bm{\alpha},
\label{3dmm}
\end{equation}
where $\vec{\bm{S}}\in \mathbb{R}^{3K\times 1}$ is the vectorized $\bm{S}$, and $\bar{\bm{S}}\in \mathbb{R}^{3K\times 1}$ is the mean shape. $\bm{U}\in \mathbb{R}^{3K\times 228}$ is the PCA basis from Basel Face Model (BFM) \cite{paysan20093d} and 3DFFA \cite{zhu2016face} for identity and expression variation, respectively. $\bar{\bm{S}}$ and $\bm{U}$ are fixed during the training and testing of CEST. With the predicted $\bm{\alpha}$, the shape $\bm{S}$ can be obtained using equation \ref{3dmm}.

\textbf{Reflectance.} Previous approaches usually predict the reflectance coefficients in a predefined model \cite{tewari2017mofa,tewari2018self}, unwrapped UV map of reflectance \cite{tran2018nonlinear,tran2019learning,lattas2020avatarme,gecer2019ganfit}, or graph representation of the reflectance \cite{wei20193d,zhou2019dense} from the image directly. In CEST, we adopt the UV map representation for reflectance. However, the prediction of the reflectance is conditioned not only on the given image $\bm{I}$, but also on the predicted viewpoint $\bm{v}$ and shape $\bm{S}$.

The process is illustrated in Fig. \ref{fig:cest_framework}.
We first compute the image-coordinate facial shape $\bm{Q} \in \mathbb{R}^{K\times 2}$ by projecting the world-coordinate facial shape $\bm{S}$ with viewpoint $\bm{v}$ using weak perspective transformation. The details of the transformation are given in Appendix \ref{tsfm}, since it is a standard formulation, and not a contribution of this paper. Next, we construct an intermediate representation, {\em i.e.} UV map of the illuminated texture $\bm{T}$ \cite{szeliski2010computer}, which is obtained by unwrapping the given image $\bm{I}$ based on the predicted face shape $\bm{Q}$. Subsequently, the UV map of reflectance $\bm{A}$ is predicted from the illuminated texture $\bm{T}$ by a reflectance function $f_r(\bm{T}; \bm{\theta}_r)$. The reflectance $\bm{R}$ can be recovered from $\bm{A}$ by UV wrapping. 
\begin{figure}[t]
  \centering
  \setlength{\abovecaptionskip}{2pt}
  \setlength{\belowcaptionskip}{-8pt}
  \renewcommand{\captionlabelfont}{\footnotesize}
  \includegraphics[width=3.25in]{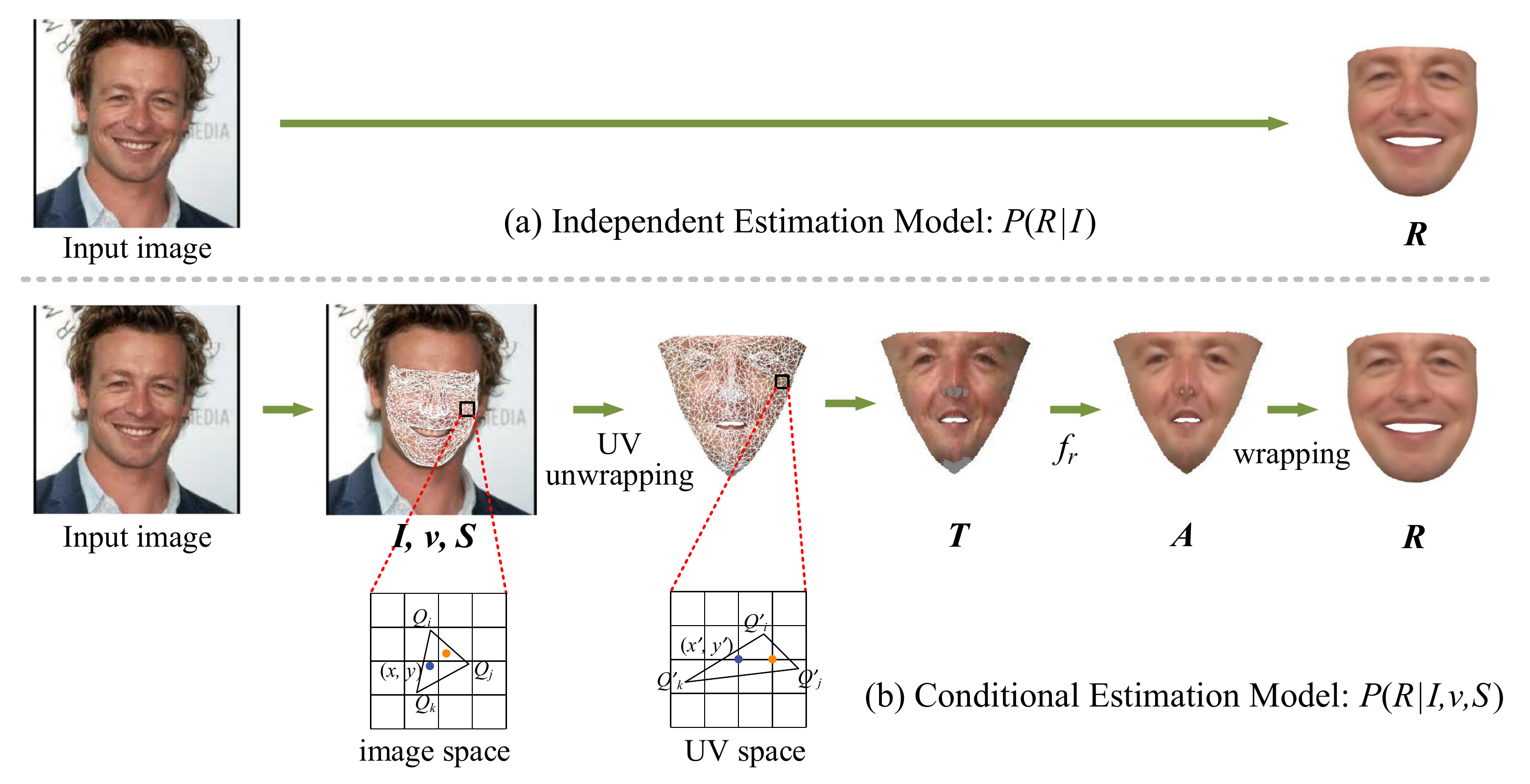}
  \caption{\footnotesize Illustration of generating the UV map of the illuminated texture.}\label{fig:cest_textureUV}
  \vspace{-0.1in}
\end{figure}

The basic idea for computing the $\bm{T}$ is illustrated in Fig. \ref{fig:cest_textureUV}. For each $\bm{T}(x', y')$ (the pixel values at position $(x', y')$), we trace its corresponding position $(x, y)$ in $\bm{I}$. The illuminated texture can be simply obtained by $\bm{T}(x', y') = \bm{I}(x, y)$, where bilinear interpolation is used for inferring the pixel values of $\bm{I}$ at position $(x, y)$ if $x$ or $y$ is not an integer. The computation of $(x, y)$ is as follows. First, the canonical face shape $\bar{\bm{S}}$ is mapped to the UV space by cylinder unwrapping. We determine the triangle enclosing the point $(x', y')$ on a grid based on the vertex connectivity, which is provided by the 3DMM. The triangle is represented by its three vertices $\bm{Q}'(i)$, $\bm{Q}'(j)$, and $\bm{Q}'(k)$. Since the topology of the facial shape in image space and UV space are the same, the vertices in these two space have one-to-one correspondence. We could easily get the corresponding vertices $\bm{Q}(i)$, $\bm{Q}(j)$, and $\bm{Q}(k)$. Now the position $(x, y)$ can be computed by $x=\kappa_1 \bm{Q}(i, 1) + \kappa_2 \bm{Q}(j, 1) + \kappa_3 \bm{Q}(k, 1)$ and $y=\kappa_1 \bm{Q}(i, 2) + \kappa_2 \bm{Q}(j, 2) + \kappa_3 \bm{Q}(k, 2)$, where the $\kappa$s are the coefficients computed by $\bm{Q}'(i)$, $\bm{Q}'(j)$, $\bm{Q}'(k)$, and $(x', y')$ in barycentric coordinate system \cite{bottema1982area}. The computation details are included in Appendix \ref{barycentric}. For the invisible triangles (caused by self-occlusion), we simply ignore them.

With the illuminated texture $\bm{T}$, the UV map of the reflectance $\bm{A}$ can be produced by a function $f_r(\bm{T}; \bm{\theta}_r)$, where $\bm{\theta}_r$ is the learnable parameters. It is worth noting that the input ($\bm{T}$) and output ($\bm{A}$) of $f_r$ are spatially aligned in UV space, so the learning process can be greatly facilitated. Subsequently, the reflectance $\bm{R}$ is obtained by a wrapping function $\bm{R} = \Psi(\bm{A})$ \cite{szeliski2010computer}, which has no learnable parameters, as shown in \ref{wrapping}.


\textbf{Illumination}. Following the previous studies \cite{genova2018unsupervised,tran2019learning}, we assume the distant smooth illumination and purely \emph{Lambertian} surface properties \cite{basri2003lambertian}. Spherical Harmonics (SH) \cite{ramamoorthi2001signal} are employed to approximate the incident radiance at a surface. We use 3 SH bands, leading to 9 SH coefficients. The illumination function is defined as $f_{\bm{\ell}}(\bm{I}, \bm{T}, \bm{A}; \bm{\theta}_{\bm{\ell}}): (\bm{I}, \bm{T}, \bm{A}) \rightarrow \bm{\ell} \in \mathbb{R}^{9\times 1}$, which takes the given image, illuminated texture map and UV map of reflectance as input, and produces the illumination parameters.

So far, the 3D face model parameters $\bm{R}$, $\bm{S}$, $\bm{v}$, and $\bm{\ell}$ are predicted, and we are able to recombine them and render the image by the expert-designed rendering module, \ie $\hat{\bm{I}} = \mathcal{R}(\bm{S}, \bm{R}, \bm{v}, \bm{\ell})$. 

\vspace{-0.8mm}
\subsection{Objectives for Self-Supervised Learning}
\vspace{-0.8mm}

The functions $f_s$, $f_r$, $f_{v}$, and $f_{\ell}$ are modelled by convolutional neural networks (CNNs) with learnable parameters $\bm{\theta}_s$, $\bm{\theta}_r$, $\bm{\theta}_{v}$, and $\bm{\theta}_{\ell}$, respectively. Since all the learning modules and expert-designed renderer are differentiable, the proposed framework is end-to-end trainable. 
The learning objective is to minimize the differences between the original image $\bm{I}$ and the rendered image $\hat{\bm{I}}$. Following the practices in previous work, the learning objective does not include the pixels in nonface region, \eg hair, sunglasses, scarf, etc. We identify if a pixel belongs to face or nonface region by a face segmentation network $f_{seg}$, which is trained on CelebAMask-HQ dataset \cite{lee2020maskgan} with the segmentation labels provided in the dataset. Once trained, $f_{seg}$ is fixed during the training and testing of CEST. We denote the effective face region as a mask $\bm{M}$, so the pixel at position $(x, y)$ is included in reconstruction if $\bm{M}(i, j)=1$, and excluded if $\bm{M}(i, j)=0$. The photometric loss can be written as
\begin{equation}
\footnotesize
\begin{aligned}
\mathcal{L}_{ph} & = \mathcal{E}(\bm{I}, \bm{S}, \bm{R}, \bm{v}, \bm{\ell}, \bm{M}) \\
& = \|\bm{M}\otimes \bm{I} - \bm{M}\otimes \hat{\bm{I}}\|_1 \\
& = \|\bm{M}\otimes \bm{I} - \bm{M}\otimes \mathcal{R}(\bm{S}, \bm{R}, \bm{v}, \bm{\ell})\|_1,
\end{aligned}
\end{equation}
where $\|\cdot\|_1$ measure the $\ell_1$ distance and $\otimes$ denotes the element-wise multiplication. However, if we simply optimize $\mathcal{L}_{ph}$, CEST will learn a degraded solution, where the reflectance $\bm{A}$ simply copies the pixel values from $\bm{T}$, and $\bm{\ell}$ yields an isotropic radiator, radiating the same intensity of radiation in all directions. In this case, CEST does not learn semantically disentangled facial parameters, but leads to a perfect reconstruction for $\hat{\bm{I}}$.

To avoid this, we adopt the symmetry and consistency constraints for reflectance. The facial reflectance is assumed to be horizontally symmetric and consistent in a video clip. Suppose $\bm{I}_i$ and $\bm{I}_j$ are two face images from the same video clip. One of the possible solutions is to add the regularization terms $\|\bm{R}_i -\bm{R}^{\Join}_i\|$, $\|\bm{R}_j -\bm{R}^{\Join}_j\|$, and $\|\bm{R}_i -\bm{R}_j\|$ to the learning objective, where $\bm{R}^{\Join}_i$ and $\bm{R}^{\Join}_j$ are the horizontally flipped versions of $\bm{R}_i$ and $\bm{R}_j$. However, it is difficult to tune loss weights to balance the reconstruction and regularization terms. Instead, we adopt an alternative solution by constructing additional reconstruction terms as constraints \cite{wu2020unsupervised}. The learning objective for reconstructing $\bm{I}_i$ and $\bm{I}_j$ can be written as
\begin{equation}
\footnotesize
\begin{aligned}
\mathcal{L}_{ph} = &\  \mathcal{E}(\bm{I}_i, \bm{S}_i, \bm{R}_i, \bm{v}_i, \bm{\ell}_i, \bm{M}_i) + \mathcal{E}(\bm{I}_j, \bm{S}_j, \bm{R}_j, \bm{v}_j, \bm{\ell}_j, \bm{M}_j) \ \\
+ & \ \mathcal{E}(\bm{I}_i, \bm{S}_i, \bm{R}_j, \bm{v}_i, \bm{\ell}_i, \bm{M}_i) + \mathcal{E}(\bm{I}_j, \bm{S}_j, \bm{R}_i, \bm{v}_j, \bm{\ell}_j, \bm{M}_j)\\
+ & \ \mathcal{E}(\bm{I}_i, \bm{S}_i, \bm{R}^{\Join}_i, \bm{v}_i, \bm{\ell}_i, \bm{M}_i) + \mathcal{E}(\bm{I}_j, \bm{S}_j, \bm{R}^{\Join}_j, \bm{v}_j, \bm{\ell}_j, \bm{M}_j)\\
+ & \ \mathcal{E}(\bm{I}_i, \bm{S}_i, \bm{R}^{\Join}_j, \bm{v}_i, \bm{\ell}_i, \bm{M}_i) + \mathcal{E}(\bm{I}_j, \bm{S}_j, \bm{R}^{\Join}_i, \bm{v}_j, \bm{\ell}_j, \bm{M}_j)
\end{aligned}
\end{equation}
\textbf{Stochastic optimization.} As can be seen, the number of reconstruction terms is increased dramatically. From $n$ frames of the same video, $2n^2$ reconstruction terms can be constructed. This is not scalable. To address this problem, we propose to optimize the learning objective in a stochastic way. For each training iteration, only a subset of the reconstruction terms are optimized. Specifically, a set of video frames $\{\bm{I}_1, \bm{I}_2, ..., \bm{I}_N\}$ are randomly sampled from different videos. The frames are grouped by videos, labeled as $\bm{\xi} = \{\xi_1, \xi_2,..., \xi_N\}$. For any $\bm{I}_i$, instead of enumerating all the possible reflectances and obtaining numerous reconstruction terms, we randomly select some other frame from the same video, denoted as $\bm{I}_j$ (under the condition of $\xi_j=\xi_i$), and use $\bm{R}_j$ and $\bm{R}^{\Join}_i$ to construct two reconstruction terms for $\bm{I}_i$. With this strategy, the number of reconstruction terms is reduced from $O(n^2)$ to $O(n)$. Formally, the learning objective can be written as

\vspace{-4mm}

\begin{equation}
\footnotesize
\begin{aligned}
\mathcal{L}_{ph} =  \frac{1}{N}\sum_{i=1, \xi_j = \xi_i}^{N}  \big( & \mathcal{E}(\bm{I}_i, \bm{S}_i, \bm{R}_j, \bm{v}_i, \bm{\ell}_i, \bm{M}_i) \\[-2mm]
& \ + \  \mathcal{E}(\bm{I}_i, \bm{S}_i, \bm{R}^{\Join}_i, \bm{v}_i, \bm{\ell}_i, \bm{M}_i) \big).
\end{aligned}
\end{equation}
\vspace{0.1mm}

To stabilize the training of CEST, we use 2D key points via $\mathcal{L}_{kp}=\frac{1}{NN_{kp}}\sum_{i=1}^{N}\sum_{j=1}^{N_{kp}}\|\bm{Q}_i(k_{j}) - \bm{q}_i(j)\|_1$ where $\bm{q}(j)$ is the set of detected 2D key points on image, and $k_j$ is the index of the vertex associating to the 2D key point. We also regularize the energies of shape coefficients with $\mathcal{L}_{rg} = \frac{1}{N}\sum_{i=1}^{N} \|\bm{\alpha}_{i}\|_2^2 $. An off-the-shelf landmark detector~\cite{bulat2017far} is used to produce $N_{kp}=68$ key points for a detected face. The total loss consists of the following terms:

\vspace{-2.5mm}

\begin{equation}
\footnotesize
\mathcal{L} =  \mathcal{L}_{ph} + \lambda_1 \mathcal{L}_{kp} + \lambda_2 \mathcal{L}_{rg}
\end{equation}
where $\lambda_1$ and $\lambda_2$ are hyperparameters.
\begin{figure*}[t]
  \centering
  \setlength{\abovecaptionskip}{3pt}
 \setlength{\belowcaptionskip}{-8pt}
  \renewcommand{\captionlabelfont}{\footnotesize}
 \includegraphics[width=6.5in]{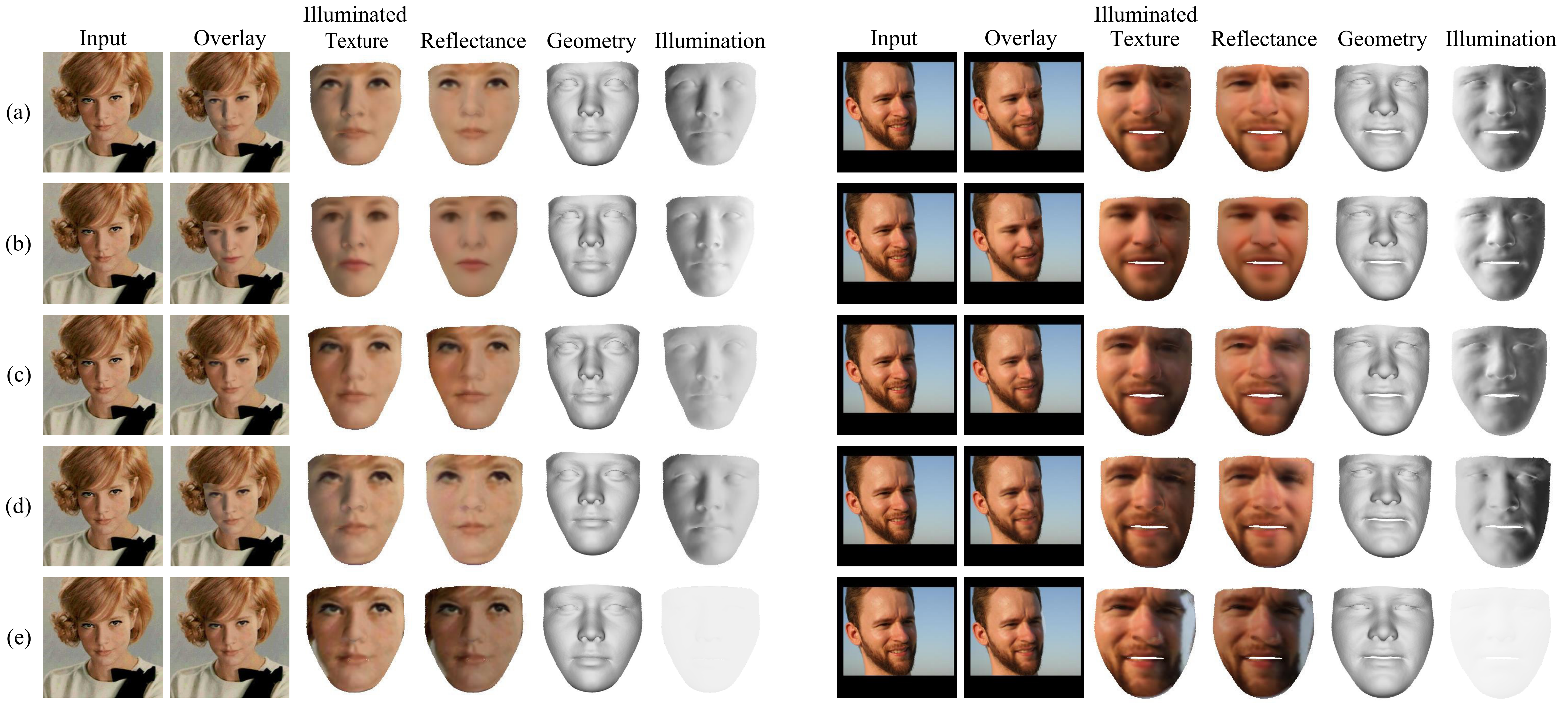}
  \caption{\footnotesize Ablations. (a) CEST with two constraints. (b) Uncoupled CEST with two constraints. (c) CEST with only reflectance consistency constraint. (d) CEST with reflectance symmetry constraint (\textbf{the number of video frames is 1}). (e) CEST with no constraint on reflectance. {\color{gray}Errata: the row (d) in the ICCV 2021 version mistakenly uses the same results as the row (c). We have fixed it in this version. Please refer to our latest arXiv version for up-to-date results.}}\label{fig:ablation}
\end{figure*}
\vspace{-0.8mm}
\section{Experiments}
\vspace{-0.8mm}
We qualitatively and quantitatively evaluate CEST with ablation experiments and comparisons to state-of-the-art methods \cite{tewari2017mofa, kim2018learning, tewari2019fml, feng2018joint}. In ablation experiments, we compare CEST to the independent version of CEST (IEST) where facial parameters are estimated in a uncoupled way, and other variants trained with different constraints. Qualitative results include the predicted shape, reflectance, illumination, reconstructed face, etc. We also show the relighted faces, which are obtained by illuminating reflectances with different illuminations. Quantitative results evaluate the qualities of the predicted shape and rendered face. The metrics we used are normalized mean error (NME) \cite{jackson2017large} and photometric error for shape and rendered face, respectively. NME is defined as the average per-vertex Euclidean distance between the predicted and targeted point clouds normalized by the outer 3D interocular distance. Photometric error is the mean absolute errors between pixel values in the original images and reconstruction images.

\vspace{-0.8mm}
\subsection{Experimental Settings}
\vspace{-0.8mm}

For fair comparison, we train two separate CEST models with VoxCeleb1 \cite{nagrani2017voxceleb} and 300W-LP \cite{zhu2016face} respectively. VoxCeleb1 is a video dataset collected from the Internet. The videos of speakers are captured in different in-the-wild scenarios. A subset of 4,727 videos of 267 persons are used in the training, leading to 6,279,609 video frames. The faces in video frames are cropped to the size of $256\times 256$ based on the detected facial key points using \cite{bulat2017far}. 300W-LP is a synthetic image dataset, containing 122,450 images provided with dense landmarks. Since we focus on self-supervised learning, we only use a sparse set of 68 sparse landmarks as a regularization in training.

\textbf{Training}. The network architectures are given in Appendix \ref{netarch}. For the training with VoxCeleb1, the minibatch consists of 128 video frames from 32 clips. For each video clip, we randomly selected 4 video frames. The training is completed at 50K iterations. For the training with 300W-LP, the minibatch consisted 128 randomly selected images, and the total iteration is 20K. For both models, we used Adam \cite{kingma2014adam} optimizer with learning rate of 0.001. $\lambda_1$ and $\lambda_2$ are $1$ and $0.1$ unless stated otherwise.

\vspace{-0.8mm}
\subsection{Ablation Experiments}
\vspace{-0.8mm}

The results of ablation study are shown in Fig. \ref{fig:ablation}. We first present the original and reconstructed image (overlay) for comparison, following by the reflectance, illuminated texture, facial shape (geometry), and illumination in canonical view. More ablations can be found in Appendix \ref{moreablation}.

\textbf{CEST and IEST.} IEST is trained with the same settings as CEST, except the facial parameters are estimated independently from image during training and testing. The results are shown in Fig. \ref{fig:ablation} (a) and (b), respectively. We can see that CEST produces realistic overlay, disentangled reflectance and illumination, and geometry with personal characteristics and expressions. Compared to CEST, IEST achieves reasonable results, but the reflectances are not as detailed as those from CEST, resulting in inferior overlays and illuminated textures. It validates our hypothesis that the coupled estimation can better formulate the problem and facilitate the learning.

\textbf{Reflectance symmetry and consistency constraints.} We train multiple variants of CEST with only symmetry constraint, only consistency constraints, and without the two constraints, and show their results in Fig.~\ref{fig:ablation} (c), (d), and (e), respectively. Compared (a) and (c) we observe that the reflectance symmetry constraint leads to better reflectance and illumination separation. This is because the horizontally flipped video frames can provide more illumination variations to the training set, enabling CEST to learn to model different illuminations properly. On the other hand, if the reflectance consistency in video clip is not used, the decomposition of reflectance and illumination is not performed well. Some illumination remains around the eyes region in the reflectance (see the right hand side of the Fig.~\ref{fig:ablation} (d)). Lastly, if we do not use any constraints on reflectance, CEST learns the degraded solution (Fig. \ref{fig:ablation}(e)), where the reflectance simply copies the pixel values from the image, and illumination is an isotropic radiator, radiating the same intensity of radiation in all directions. Moreover, we note that the degraded solution also affects the learned facial shape, which has less personal characteristics in Fig.~\ref{fig:ablation} (e).

\subsection{Qualitative Results}
In this section, we compare CEST to most relevant state-of-art methods with qualitative results. More qualitative results are included in Appendix \ref{morequalitative}.

\begin{figure}[t]
  \centering
  \setlength{\abovecaptionskip}{3pt}
 \setlength{\belowcaptionskip}{-8pt}
  \renewcommand{\captionlabelfont}{\footnotesize}
 \includegraphics[width=3.2in]{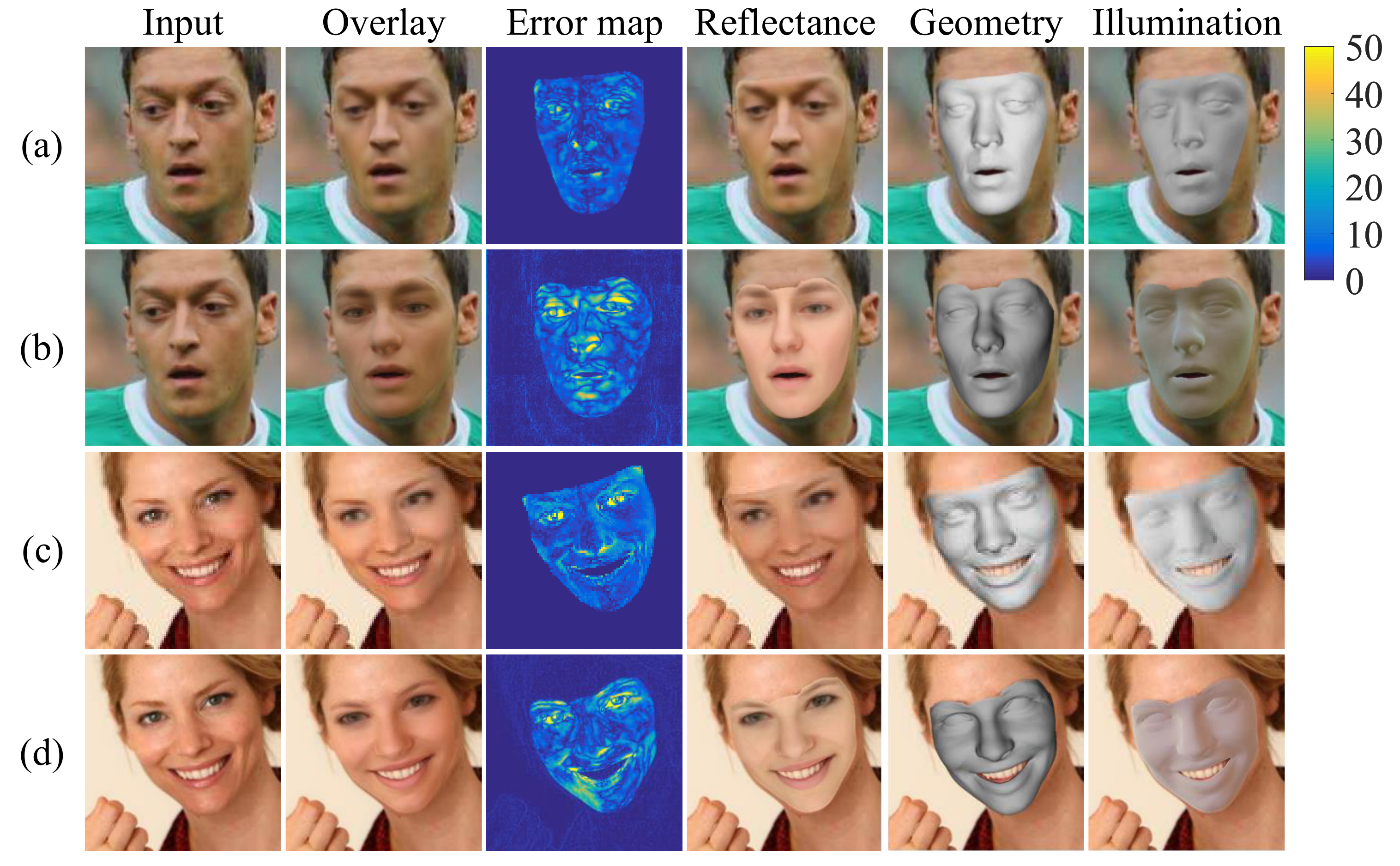}
  \caption{\footnotesize Comparisons with MoFA. (a) and (c) are results from CEST. (b) and (d) are results from MoFA. Images are from CelebA dataset \cite{liu2015deep}}\label{fig:mofa}
\end{figure}
\textbf{Comparison to MoFA \cite{tewari2017mofa}.} MoFA is a fully model-based framework. Its representation power is limited by the linear 3DMM model. In addition, all facial parameters from MoFA are independently predicted from the original image. On the contrary, we use a model-free method for reflectance, and the whole inference process is based on coupled estimation. We visualize the overlay, reflectance, geometry, illumination , as well as the errors between input and rendered image (overlays) in Fig.~\ref{fig:mofa}. As can be observed, results from MoFA suffer from out-of-subspace reflectance variations. Compared to MoFA, we obtain comparable shape, but significantly better reflectance, illumination, and rendered face by capturing more details. 

\begin{figure}[t]
  \centering
  \setlength{\abovecaptionskip}{2pt}
 \setlength{\belowcaptionskip}{-7pt}
  \renewcommand{\captionlabelfont}{\footnotesize}
 \includegraphics[width=3.2in]{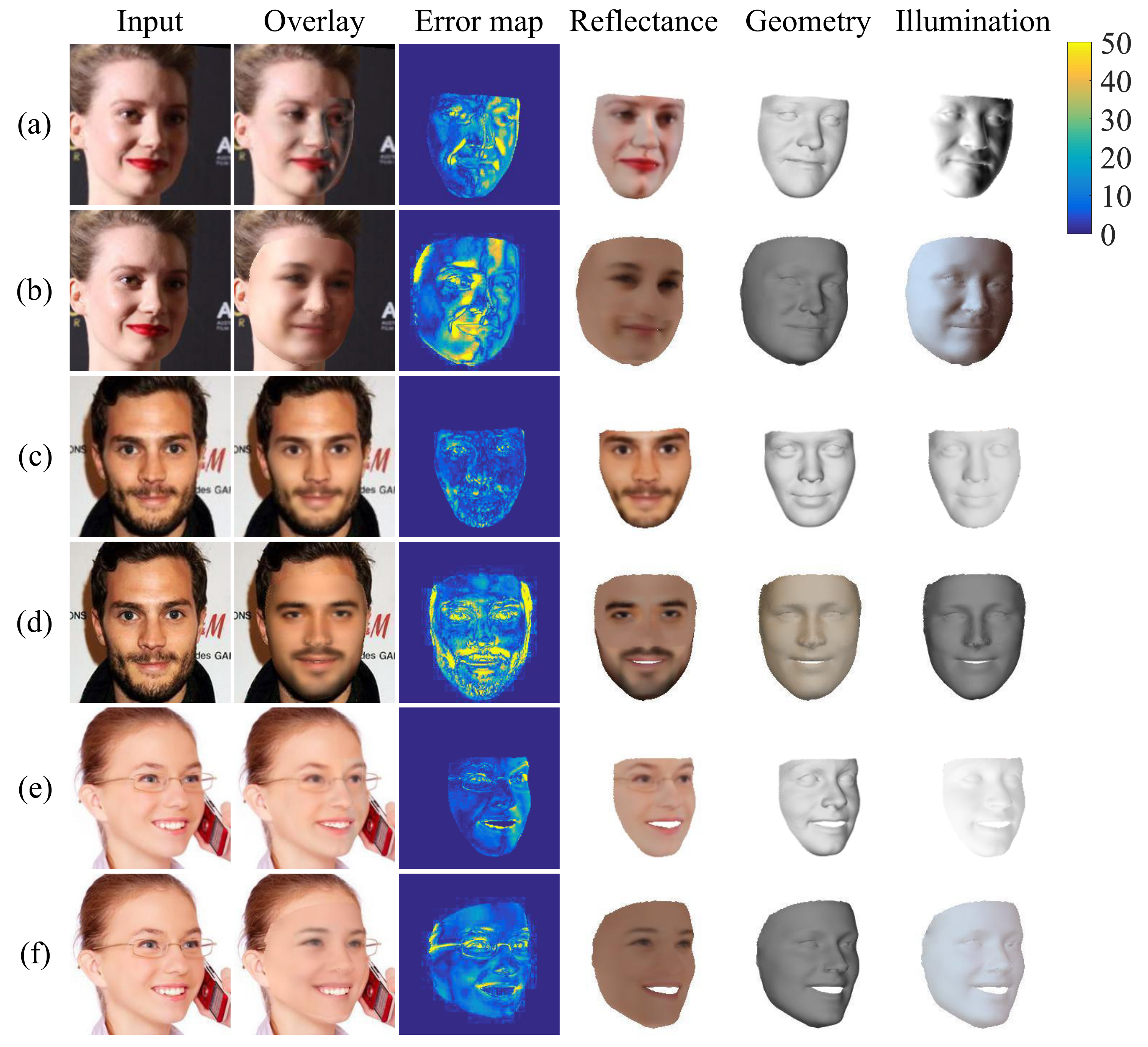}
  \caption{\footnotesize Comparisons with nonlinear 3DMM. (a), (c), and (e) are results from CEST. (b), (d), and (f) are results from N3DMM. Images are from AFLW2000-3D dataset \cite{zhu2016face}}\label{fig:n3dmm}
\end{figure}
\textbf{Comparison to N3DMM \cite{tran2019learning}.} N3DMM generalizes 3DMM model to a nonlinear space and improves the quality of rendered faces. However, N3DMM also infers the reflectance from the input image only, and uses too many heuristic constraints, e.g. reflectance constancy, shape smoothness, supervised pretraining, etc. So their models can only capture low-frequency variations on reflectance. For example, in Fig. \ref{fig:n3dmm} (b) the lip stick is missing in the reflectance, and the skin colors in reflectances are almost identical for different persons. These limitations lead to higher reconstruction error. In contrast, our results produce realistic reconstruction, with more accurate reflectance and illumination, as well as lower reconstructed error (Fig.~\ref{fig:n3dmm}). 

\begin{figure}[t]
  \centering
  \setlength{\abovecaptionskip}{5pt}
 \setlength{\belowcaptionskip}{-5pt}
  \renewcommand{\captionlabelfont}{\footnotesize}
 \includegraphics[width=3.2in]{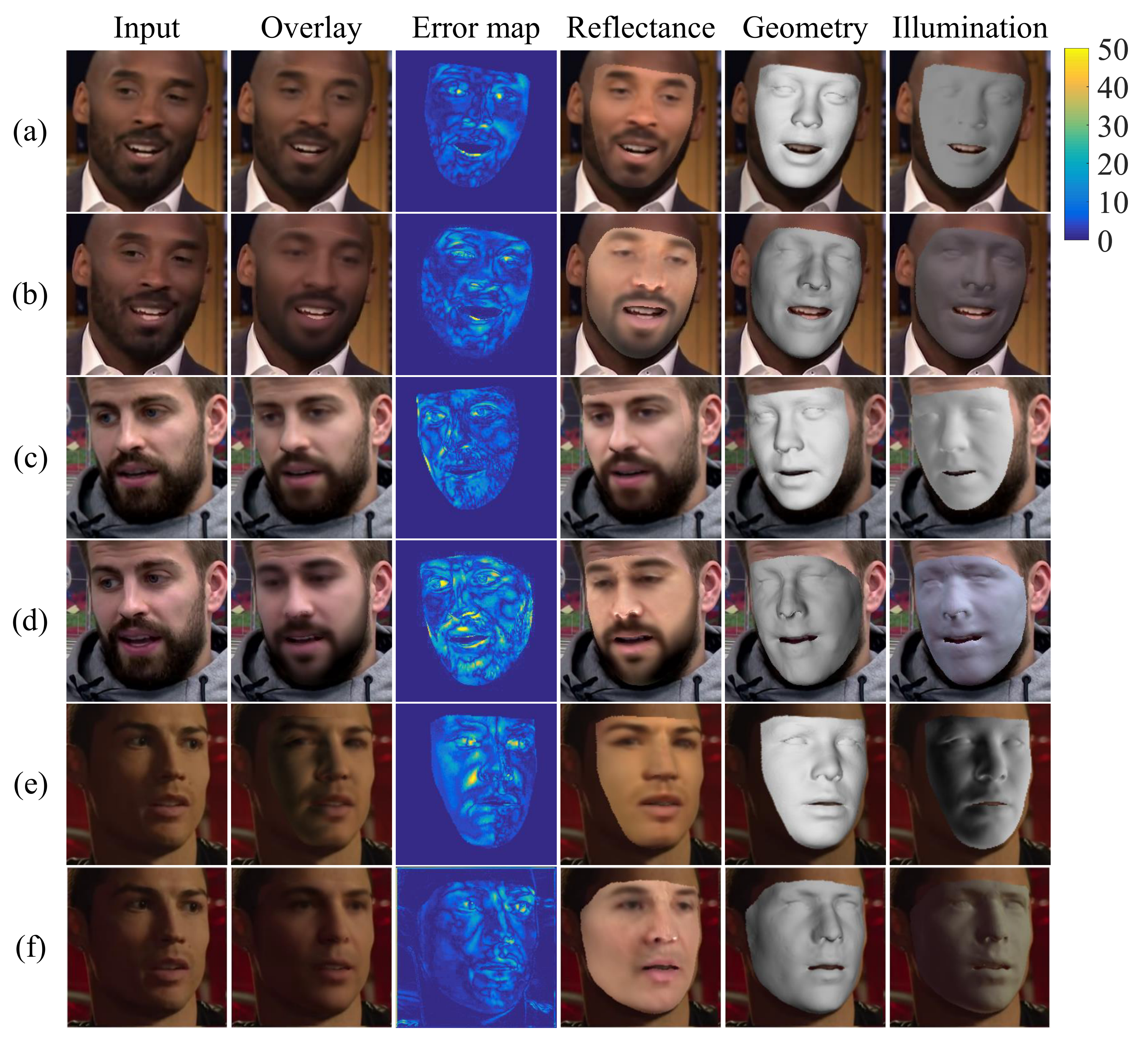}
  \caption{\footnotesize Comparisons with FML. (a), (c), and (e) are results from CEST. (b), (d), and (f) are results from FML. Images are from the video frames in VoxCeleb1 dataset \cite{nagrani2017voxceleb}}\label{fig:fml}
\end{figure}
\textbf{Comparison to FML \cite{tewari2019fml}.} FML properly incorporates video clues in training and can render realistic faces. However, its reconstructed reflectances are prone to an average skin color. In comparison, CEST yields more accurate skin color (see Fig.~\ref{fig:fml} (a), (c), and (e)) by incorporating the learned shape and viewpoint in the estimation of reflectance. Qualitative results clearly show that our results have more reasonable disentanglement between reflectance and illumination. They also contribute to better visual quality of rendered faces. Notably, there are considerable differences in the eye and nose regions from the overlay in Fig.~\ref{fig:fml}.

\begin{figure}[t]
  \centering
  \setlength{\abovecaptionskip}{3pt}
 \setlength{\belowcaptionskip}{-8pt}
  \renewcommand{\captionlabelfont}{\footnotesize}
 \includegraphics[width=3.0in]{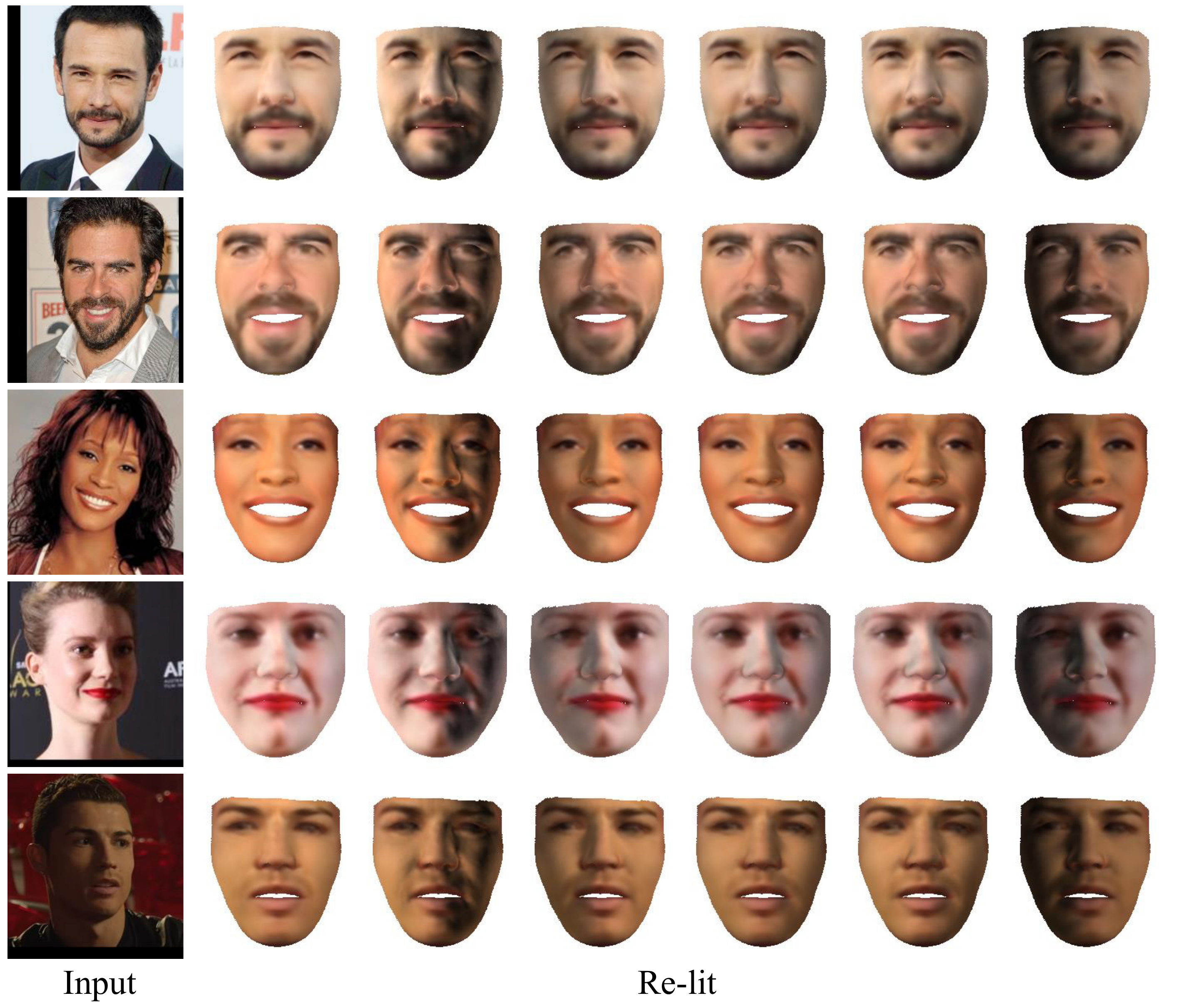}
  \caption{\footnotesize Lighting transfer results.}\label{fig:relighting}
\end{figure}
\textbf{Relighting.} Since CEST predicts the reflectances of faces, they can be easily re-lighted with different lighting conditions. Fig. \ref{fig:relighting} shows the re-lit faces in canonical view. In particular, the last two target faces are under harsh lighting, which also examines the illumination removal ability of CEST. The re-lit results again validate that CEST is capable of estimating well-disentangled facial parameters and capturing the reflectance and illumination variations in real-world face images.

\vspace{-0.8mm}
\subsection{Quantitative Results}
\vspace{-0.8mm}

We first perform quantitative evaluations on the AFLW2000-3D dataset, including 2,000 unconstrained face images with large pose variations. The ground truth of AFLW2000-3D is given by the results from 3DMM fitting, which may be somewhat noisy. The second evaluation is on MICC Florence 3D Face dataset, which consists of high-resolution 3D scans from 53 subjects. We follow the practices in \cite{jackson2017large} to render 2,550 testing images using the provided 3D scans. Each subject is rendered in 20 difference poses using a pitch of -15, 20 or 25 degrees and a yaw of -80, -40, 0, 40 or 80 degrees.

\begin{figure}[t]
  \centering
  \setlength{\abovecaptionskip}{5pt}
  \setlength{\belowcaptionskip}{-10pt}
    \renewcommand{\captionlabelfont}{\footnotesize}
 \includegraphics[width=3.25in]{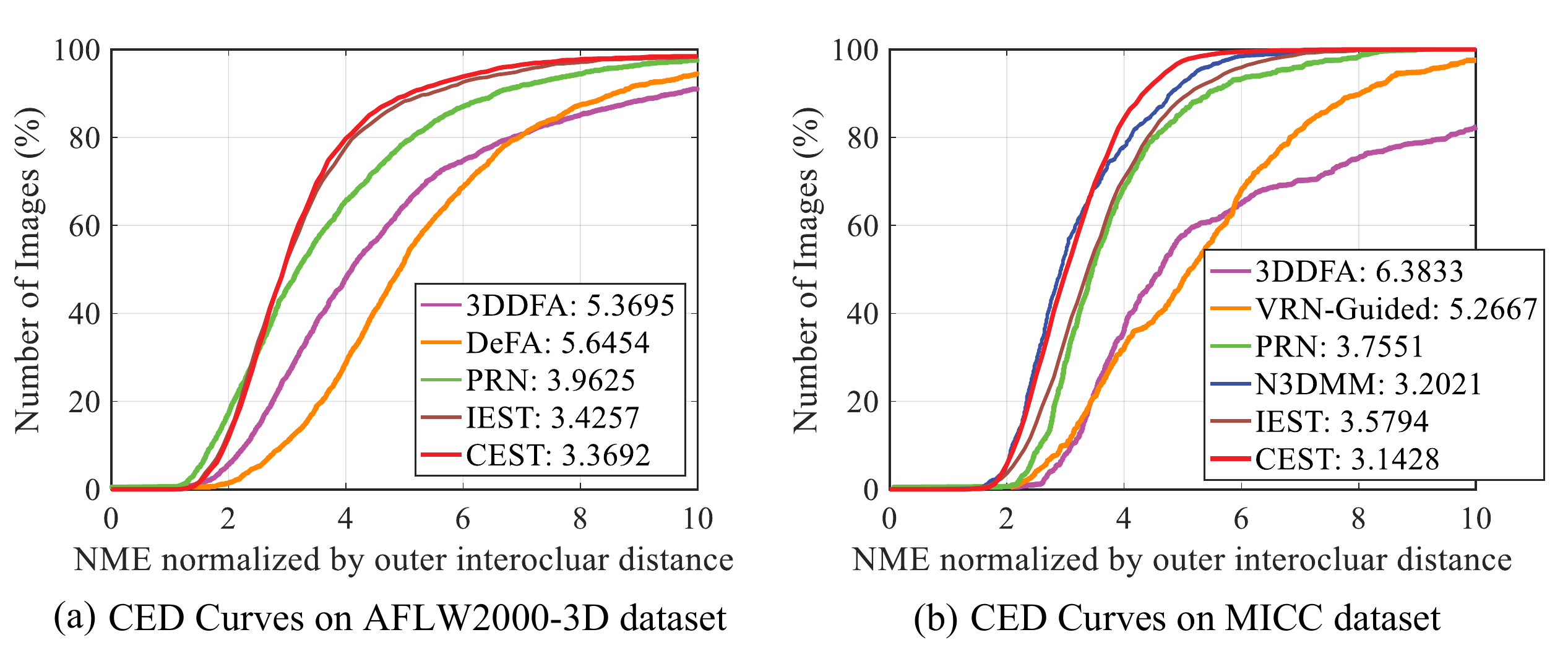}
  \caption{\footnotesize CED curves on AFLW2000-3D and MICC datasets. For example, a point at (4, 63) means 63\% of images have NME less than 4.}\label{fig:quantative}
\end{figure}

In order to compare with previous work, NME is computed based on a set of 19,618 vertices defined by \cite{jackson2017large} in their evaluation. The point correspondences are determined by the iterative closest point (ICP) algorithm \cite{besl1992method}. We compute the cumulative errors distribution (CED) curves and compare it to current prevailing methods such as 3DDFA \cite{zhu2016face}, DeFA \cite{liu2017dense}, and PRN \cite{feng2018joint} on AFLW2000-3D. For MICC, we compare CEST to 3DDFA \cite{zhu2016face}, VRN \cite{jackson2017large}, and PRN \cite{feng2018joint}. The results are given in Fig.~\ref{fig:quantative}. CEST achieves 3.37 and 3.14 NME on AFLW2000-3D and MICC datasets, respectively. More interestingly, our method performs better than the {\em fully supervised} techniques for shape estimation, e.g. 3DDFA (5.37 on AFLW2000-3D and 6.38 on MICC) and PRN (3.96 on AFLW2000-3D and 3.76 on MICC). Additionally, our method can also estimate facial reflectance and illumination, while both 3DDFA and PRN can not. Compared to N3DMM on MICC dataset, CEST achieves slightly lower NME (3.14 vs. 3.20). Notably, N3DMM uses dense landmarks for supervised pretraining while CEST only uses the 68 sparse landmarks. More quantitative comparisons can be found in Appendix~\ref{morequantitative}. 

\vspace{-1mm}
\section{Conclusion and Future Work}
\vspace{-0.7mm}
We have proposed a conditional estimation framework, called CEST, for 3D face reconstruction from single-view images. CEST addresses the reconstruction problem with a more general formulation, which does not assume conditional independence. We have also proposed a specific decomposition for the conditional probability of different 3D facial parameters. Together with the reflectance symmetry and consistency constraints, CEST can be trained efficiently with video datasets. Both qualitative and quantitative results prove that the conditional estimation is useful. CEST is able to produce high quality and well-disentangled facial parameters for single-view images. 

The proposed CEST can be improved from many aspects. Firstly, more accurate and unambiguous facial parameters can be obtained by exploring the temporal information in video. Second, the performance of shape estimation can be boosted by a more advanced morphable model, which also benefits the subsequent estimations of other facial parameters. Moreover, adding perceptual loss could also be an effective way to improve the visual quality of the facial parameters.

{\small
\bibliographystyle{ieee_fullname}
\bibliography{egbib}

\begin{thebibliography}{10}\itemsep=-1pt

\bibitem{bagdanov2011florence}
Andrew~D Bagdanov, Alberto Del~Bimbo, and Iacopo Masi.
\newblock The florence 2d/3d hybrid face dataset.
\newblock In {\em Proceedings of the 2011 joint ACM workshop on Human gesture
  and behavior understanding}, pages 79--80, 2011.

\bibitem{basri2003lambertian}
Ronen Basri and David~W Jacobs.
\newblock Lambertian reflectance and linear subspaces.
\newblock {\em IEEE transactions on pattern analysis and machine intelligence},
  25(2):218--233, 2003.

\bibitem{besl1992method}
Paul~J Besl and Neil~D McKay.
\newblock Method for registration of 3-d shapes.
\newblock In {\em Sensor fusion IV: control paradigms and data structures},
  volume 1611, pages 586--606. International Society for Optics and Photonics,
  1992.

\bibitem{blanz1999morphable}
Volker Blanz and Thomas Vetter.
\newblock A morphable model for the synthesis of 3d faces.
\newblock In {\em Proceedings of the 26th annual conference on Computer
  graphics and interactive techniques}, pages 187--194, 1999.

\bibitem{booth20173d}
James Booth, Epameinondas Antonakos, Stylianos Ploumpis, George Trigeorgis,
  Yannis Panagakis, and Stefanos Zafeiriou.
\newblock 3d face morphable models" in-the-wild".
\newblock In {\em 2017 IEEE Conference on Computer Vision and Pattern
  Recognition (CVPR)}, pages 5464--5473. IEEE, 2017.

\bibitem{bottema1982area}
O Bottema.
\newblock On the area of a triangle in barycentric coordinates.
\newblock {\em Crux Mathematicorum}, 8(8):228--231, 1982.

\bibitem{bulat2017far}
Adrian Bulat and Georgios Tzimiropoulos.
\newblock How far are we from solving the 2d \& 3d face alignment problem? (and
  a dataset of 230,000 3d facial landmarks).
\newblock In {\em International Conference on Computer Vision}, 2017.

\bibitem{cao2013facewarehouse}
Chen Cao, Yanlin Weng, Shun Zhou, Yiying Tong, and Kun Zhou.
\newblock Facewarehouse: A 3d facial expression database for visual computing.
\newblock {\em IEEE Transactions on Visualization and Computer Graphics},
  20(3):413--425, 2013.

\bibitem{feng2018joint}
Yao Feng, Fan Wu, Xiaohu Shao, Yanfeng Wang, and Xi Zhou.
\newblock Joint 3d face reconstruction and dense alignment with position map
  regression network.
\newblock In {\em Proceedings of the European Conference on Computer Vision
  (ECCV)}, pages 534--551, 2018.

\bibitem{garg2013dense}
Ravi Garg, Anastasios Roussos, and Lourdes Agapito.
\newblock Dense variational reconstruction of non-rigid surfaces from monocular
  video.
\newblock In {\em Proceedings of the IEEE Conference on computer vision and
  pattern recognition}, pages 1272--1279, 2013.

\bibitem{garrido2016reconstruction}
Pablo Garrido, Michael Zollh{\"o}fer, Dan Casas, Levi Valgaerts, Kiran
  Varanasi, Patrick P{\'e}rez, and Christian Theobalt.
\newblock Reconstruction of personalized 3d face rigs from monocular video.
\newblock {\em ACM Transactions on Graphics (TOG)}, 35(3):1--15, 2016.

\bibitem{garrido2016corrective}
Pablo Garrido, Michael Zollh{\"o}fer, Chenglei Wu, Derek Bradley, Patrick
  P{\'e}rez, Thabo Beeler, and Christian Theobalt.
\newblock Corrective 3d reconstruction of lips from monocular video.
\newblock {\em ACM Transactions on Graphics (TOG)}, 35(6):1--11, 2016.

\bibitem{gecer2019ganfit}
Baris Gecer, Stylianos Ploumpis, Irene Kotsia, and Stefanos Zafeiriou.
\newblock Ganfit: Generative adversarial network fitting for high fidelity 3d
  face reconstruction.
\newblock In {\em Proceedings of the IEEE Conference on Computer Vision and
  Pattern Recognition}, pages 1155--1164, 2019.

\bibitem{genova2018unsupervised}
Kyle Genova, Forrester Cole, Aaron Maschinot, Aaron Sarna, Daniel Vlasic, and
  William~T Freeman.
\newblock Unsupervised training for 3d morphable model regression.
\newblock In {\em Proceedings of the IEEE Conference on Computer Vision and
  Pattern Recognition}, pages 8377--8386, 2018.

\bibitem{ioffe2015batch}
Sergey Ioffe and Christian Szegedy.
\newblock Batch normalization: Accelerating deep network training by reducing
  internal covariate shift.
\newblock {\em arXiv preprint arXiv:1502.03167}, 2015.

\bibitem{jackson2017large}
Aaron~S Jackson, Adrian Bulat, Vasileios Argyriou, and Georgios Tzimiropoulos.
\newblock Large pose 3d face reconstruction from a single image via direct
  volumetric cnn regression.
\newblock In {\em Proceedings of the IEEE International Conference on Computer
  Vision}, pages 1031--1039, 2017.

\bibitem{kemelmacher20103d}
Ira Kemelmacher-Shlizerman and Ronen Basri.
\newblock 3d face reconstruction from a single image using a single reference
  face shape.
\newblock {\em IEEE transactions on pattern analysis and machine intelligence},
  33(2):394--405, 2010.

\bibitem{kemelmacher2011face}
Ira Kemelmacher-Shlizerman and Steven~M Seitz.
\newblock Face reconstruction in the wild.
\newblock In {\em 2011 International Conference on Computer Vision}, pages
  1746--1753. IEEE, 2011.

\bibitem{kim2018learning}
Changil Kim, Hijung~Valentina Shin, Tae-Hyun Oh, Alexandre Kaspar, Mohamed
  Elgharib, and Wojciech Matusik.
\newblock On learning associations of faces and voices.
\newblock In {\em Asian Conference on Computer Vision}, pages 276--292.
  Springer, 2018.

\bibitem{kim2018inversefacenet}
Hyeongwoo Kim, Michael Zollh{\"o}fer, Ayush Tewari, Justus Thies, Christian
  Richardt, and Christian Theobalt.
\newblock Inversefacenet: Deep monocular inverse face rendering.
\newblock In {\em Proceedings of the IEEE Conference on Computer Vision and
  Pattern Recognition}, pages 4625--4634, 2018.

\bibitem{kingma2014adam}
Diederik~P Kingma and Jimmy Ba.
\newblock Adam: A method for stochastic optimization.
\newblock {\em arXiv preprint arXiv:1412.6980}, 2014.

\bibitem{lattas2020avatarme}
Alexandros Lattas, Stylianos Moschoglou, Baris Gecer, Stylianos Ploumpis,
  Vasileios Triantafyllou, Abhijeet Ghosh, and Stefanos Zafeiriou.
\newblock Avatarme: Realistically renderable 3d facial reconstruction.
\newblock In {\em Proceedings of the IEEE/CVF Conference on Computer Vision and
  Pattern Recognition}, pages 760--769, 2020.

\bibitem{lee2020maskgan}
Cheng-Han Lee, Ziwei Liu, Lingyun Wu, and Ping Luo.
\newblock Maskgan: Towards diverse and interactive facial image manipulation.
\newblock In {\em Proceedings of the IEEE/CVF Conference on Computer Vision and
  Pattern Recognition}, pages 5549--5558, 2020.

\bibitem{liu2017dense}
Yaojie Liu, Amin Jourabloo, William Ren, and Xiaoming Liu.
\newblock Dense face alignment.
\newblock In {\em Proceedings of the IEEE International Conference on Computer
  Vision Workshops}, pages 1619--1628, 2017.

\bibitem{liu2015deep}
Ziwei Liu, Ping Luo, Xiaogang Wang, and Xiaoou Tang.
\newblock Deep learning face attributes in the wild.
\newblock In {\em ICCV}, 2015.

\bibitem{nagrani2017voxceleb}
Arsha Nagrani, Joon~Son Chung, and Andrew Zisserman.
\newblock Voxceleb: a large-scale speaker identification dataset.
\newblock {\em arXiv preprint arXiv:1706.08612}, 2017.

\bibitem{paysan20093d}
Pascal Paysan, Reinhard Knothe, Brian Amberg, Sami Romdhani, and Thomas Vetter.
\newblock A 3d face model for pose and illumination invariant face recognition.
\newblock In {\em 2009 Sixth IEEE International Conference on Advanced Video
  and Signal Based Surveillance}, pages 296--301. Ieee, 2009.

\bibitem{ramamoorthi2001signal}
Ravi Ramamoorthi and Pat Hanrahan.
\newblock A signal-processing framework for inverse rendering.
\newblock In {\em Proceedings of the 28th annual conference on Computer
  graphics and interactive techniques}, pages 117--128, 2001.

\bibitem{richardson2017learning}
Elad Richardson, Matan Sela, Roy Or-El, and Ron Kimmel.
\newblock Learning detailed face reconstruction from a single image.
\newblock In {\em Proceedings of the IEEE Conference on Computer Vision and
  Pattern Recognition}, pages 1259--1268, 2017.

\bibitem{ronneberger2015u}
Olaf Ronneberger, Philipp Fischer, and Thomas Brox.
\newblock U-net: Convolutional networks for biomedical image segmentation.
\newblock In {\em International Conference on Medical image computing and
  computer-assisted intervention}, pages 234--241. Springer, 2015.

\bibitem{sanyal2019learning}
Soubhik Sanyal, Timo Bolkart, Haiwen Feng, and Michael~J Black.
\newblock Learning to regress 3d face shape and expression from an image
  without 3d supervision.
\newblock In {\em Proceedings of the IEEE/CVF Conference on Computer Vision and
  Pattern Recognition}, pages 7763--7772, 2019.

\bibitem{sela2017unrestricted}
Matan Sela, Elad Richardson, and Ron Kimmel.
\newblock Unrestricted facial geometry reconstruction using image-to-image
  translation.
\newblock In {\em Proceedings of the IEEE International Conference on Computer
  Vision}, pages 1576--1585, 2017.

\bibitem{sengupta2018sfsnet}
Soumyadip Sengupta, Angjoo Kanazawa, Carlos~D Castillo, and David~W Jacobs.
\newblock Sfsnet: Learning shape, reflectance and illuminance of facesin the
  wild'.
\newblock In {\em Proceedings of the IEEE Conference on Computer Vision and
  Pattern Recognition}, pages 6296--6305, 2018.

\bibitem{shang2020self}
Jiaxiang Shang, Tianwei Shen, Shiwei Li, Lei Zhou, Mingmin Zhen, Tian Fang, and
  Long Quan.
\newblock Self-supervised monocular 3d face reconstruction by occlusion-aware
  multi-view geometry consistency.
\newblock In {\em Computer Vision--ECCV 2020: 16th European Conference,
  Glasgow, UK, August 23--28, 2020, Proceedings, Part XV 16}, pages 53--70.
  Springer, 2020.

\bibitem{shi2014automatic}
Fuhao Shi, Hsiang-Tao Wu, Xin Tong, and Jinxiang Chai.
\newblock Automatic acquisition of high-fidelity facial performances using
  monocular videos.
\newblock {\em ACM Transactions on Graphics (TOG)}, 33(6):1--13, 2014.

\bibitem{szeliski2010computer}
Richard Szeliski.
\newblock {\em Computer vision: algorithms and applications}.
\newblock Springer Science \& Business Media, 2010.

\bibitem{tewari2019fml}
Ayush Tewari, Florian Bernard, Pablo Garrido, Gaurav Bharaj, Mohamed Elgharib,
  Hans-Peter Seidel, Patrick P{\'e}rez, Michael Zollhofer, and Christian
  Theobalt.
\newblock Fml: Face model learning from videos.
\newblock In {\em Proceedings of the IEEE Conference on Computer Vision and
  Pattern Recognition}, pages 10812--10822, 2019.

\bibitem{tewari2018self}
Ayush Tewari, Michael Zollh{\"o}fer, Pablo Garrido, Florian Bernard, Hyeongwoo
  Kim, Patrick P{\'e}rez, and Christian Theobalt.
\newblock Self-supervised multi-level face model learning for monocular
  reconstruction at over 250 hz.
\newblock In {\em Proceedings of the IEEE Conference on Computer Vision and
  Pattern Recognition}, pages 2549--2559, 2018.

\bibitem{tewari2017mofa}
Ayush Tewari, Michael Zollhofer, Hyeongwoo Kim, Pablo Garrido, Florian Bernard,
  Patrick Perez, and Christian Theobalt.
\newblock Mofa: Model-based deep convolutional face autoencoder for
  unsupervised monocular reconstruction.
\newblock In {\em Proceedings of the IEEE International Conference on Computer
  Vision Workshops}, pages 1274--1283, 2017.

\bibitem{thies2016face2face}
Justus Thies, Michael Zollhofer, Marc Stamminger, Christian Theobalt, and
  Matthias Nie{\ss}ner.
\newblock Face2face: Real-time face capture and reenactment of rgb videos.
\newblock In {\em Proceedings of the IEEE conference on computer vision and
  pattern recognition}, pages 2387--2395, 2016.

\bibitem{tran2018nonlinear}
Luan Tran and Xiaoming Liu.
\newblock Nonlinear 3d face morphable model.
\newblock In {\em Proceedings of the IEEE conference on computer vision and
  pattern recognition}, pages 7346--7355, 2018.

\bibitem{tran2019learning}
Luan Tran and Xiaoming Liu.
\newblock On learning 3d face morphable model from in-the-wild images.
\newblock {\em IEEE transactions on pattern analysis and machine intelligence},
  2019.

\bibitem{tuan2017regressing}
Anh Tuan~Tran, Tal Hassner, Iacopo Masi, and G{\'e}rard Medioni.
\newblock Regressing robust and discriminative 3d morphable models with a very
  deep neural network.
\newblock In {\em Proceedings of the IEEE conference on computer vision and
  pattern recognition}, pages 5163--5172, 2017.

\bibitem{wei20193d}
Huawei Wei, Shuang Liang, and Yichen Wei.
\newblock 3d dense face alignment via graph convolution networks.
\newblock {\em arXiv preprint arXiv:1904.05562}, 2019.

\bibitem{wu2020unsupervised}
Shangzhe Wu, Christian Rupprecht, and Andrea Vedaldi.
\newblock Unsupervised learning of probably symmetric deformable 3d objects
  from images in the wild.
\newblock In {\em Proceedings of the IEEE/CVF Conference on Computer Vision and
  Pattern Recognition}, pages 1--10, 2020.

\bibitem{yin20063d}
Lijun Yin, Xiaozhou Wei, Yi Sun, Jun Wang, and Matthew~J Rosato.
\newblock A 3d facial expression database for facial behavior research.
\newblock In {\em 7th international conference on automatic face and gesture
  recognition (FGR06)}, pages 211--216. IEEE, 2006.

\bibitem{zhang2013high}
Xing Zhang, Lijun Yin, Jeffrey~F Cohn, Shaun Canavan, Michael Reale, Andy
  Horowitz, and Peng Liu.
\newblock A high-resolution spontaneous 3d dynamic facial expression database.
\newblock In {\em 2013 10th IEEE International Conference and Workshops on
  Automatic Face and Gesture Recognition (FG)}, pages 1--6. IEEE, 2013.

\bibitem{zhou2019dense}
Yuxiang Zhou, Jiankang Deng, Irene Kotsia, and Stefanos Zafeiriou.
\newblock Dense 3d face decoding over 2500fps: Joint texture \& shape
  convolutional mesh decoders.
\newblock In {\em Proceedings of the IEEE Conference on Computer Vision and
  Pattern Recognition}, pages 1097--1106, 2019.

\bibitem{zhu2016face}
Xiangyu Zhu, Zhen Lei, Xiaoming Liu, Hailin Shi, and Stan~Z Li.
\newblock Face alignment across large poses: A 3d solution.
\newblock In {\em Proceedings of the IEEE conference on computer vision and
  pattern recognition}, pages 146--155, 2016.

\end{thebibliography}
}
\newpage

\onecolumn
\begin{appendix}
\begin{center}

{\LARGE \textbf{Appendix}}
\vspace{2mm}

\end{center}

\section{Approach}
\subsection{Image Cropping}
\label{cropping}
The viewpoint $\bm{v}$ comprises the scale factor $\bm{v}_1$, 3D spatial rotation parameters $[\bm{v}_2, \bm{v}_3, \bm{v}_4]$, and 3D translation parameters $[\bm{v}_5, \bm{v}_6, \bm{v}_7]$. The original image $\bm{I}$ is cropped to its canonical view in 2D plane with viewpoint $v$. The cropping is given by $(\bm{I}\circ \bm{v})(x', y') = \bm{I}(x, y)$, where the transformation from $(x', y')$ to $(x, y)$ is formulated in the following.
\begin{equation}
\footnotesize
\begin{split}
\left [\begin{array}{c}
x\\
y\\
\end{array} \right] & = \left[ \begin{array}{ccc}
 \exp(\bm{v}_1)\cdot\cos \bm{v}_4  &  \exp(\bm{v}_1)\cdot\sin \bm{v}_4 & \bm{v}_5 \\
 -\exp(\bm{v}_1)\cdot\sin \bm{v}_4  &  \exp(\bm{v}_1)\cdot\cos \bm{v}_4 & \bm{v}_6  \\
\end{array} \right] \left [\begin{array}{c}
x'\\
y'\\
\end{array} \right]
\end{split}
\end{equation}
Bilinear interpolation is used if $x$ or $y$ is not an integer.

\subsection{Weak Perspective Transformation}
\label{tsfm}
The 3D spatial rotation is represented by a rotation vector $\bm{w} = [\bm{v}_2; \bm{v}_3; \bm{v}_4] \in \mathbb{R}^{3 \times 1}$: the unit vector $\bm{u} = \frac{\bm{w}}{\|\bm{w}\|_2}$ is the axis of rotation, and the magnitude $\phi = \|\bm{w}\|_2$ is the rotation angle. The weak perspective transformation is used to project the world-coordinate facial shape $\bm{S}$ to image-coordinate $\bm{Q}$, as formulated in
\begin{equation}
\footnotesize
\left [\begin{array}{c}
 \bm{Q}(i, 1)   \\
 \bm{Q}(i, 2)   \\
 \bm{Q}(i, 3)   \\
\end{array} \right] = \exp(\bm{v}_1)\cdot \Bigg( \bm{w}\bm{w}^{\intercal}\left [\begin{array}{c}
 \bm{S}(i, 1)  \\
 \bm{S}(i, 2)  \\
 \bm{S}(i, 3)  \\
\end{array} \right] 
+ \ (\cos \phi) \cdot (1 - \bm{w}\bm{w}^{\intercal}) \left [\begin{array}{c}
 \bm{S}(i, 1)  \\
 \bm{S}(i, 2)  \\
 \bm{S}(i, 3)  \\
\end{array} \right] 
+ \ (\sin \phi) \cdot \bm{w} \times \left [\begin{array}{c}
 \bm{S}(i, 1)  \\
 \bm{S}(i, 2)  \\
 \bm{S}(i, 3)  \\
\end{array} \right] \Bigg) + \left [\begin{array}{c}
 \bm{v}_5  \\
 \bm{v}_6  \\
 \bm{v}_7  \\
\end{array} \right].
\end{equation}

\subsection{Barycentric Coefficients}
\label{barycentric}
Given the vertices of a triangle ($\bm{Q}(i),\bm{Q}(j),\bm{Q}(k) $) and its enclosing grid point $(x, y)$ on image. The barycentric coefficients can be computed by
\begin{equation}
\footnotesize
\begin{gathered}
\bm{d}_i = \left [\begin{array}{c}
\bm{Q}(j, 1) - \bm{Q}(i, 1) \\
\bm{Q}(j, 2) - \bm{Q}(i, 2) \\
\end{array} \right], \ \ \ \ \ 
\bm{d}_j = \left [\begin{array}{c}
\bm{Q}(k, 1) - \bm{Q}(i, 1) \\
\bm{Q}(k, 2) - \bm{Q}(i, 2) \\
\end{array} \right], \ \ \ \ \ 
\bm{d}_k = \left [\begin{array}{c}
x - \bm{Q}(i, 1) \\
y - \bm{Q}(i, 2) \\
\end{array} \right], \\
d_{ii} = \bm{d}_{i}^\intercal\bm{d}_{i}, \ \ \ \ \  
d_{jj} = \bm{d}_{j}^\intercal\bm{d}_{j}, \ \ \ \ \  d_{ij} = \bm{d}_{i}^\intercal\bm{d}_{j}, \ \ \ \ \  d_{ki} = \bm{d}_{k}^\intercal\bm{d}_{i}, \ \ \ \ \  d_{kj} = \bm{d}_{k}^\intercal\bm{d}_{i}, \\
\kappa_2 = \frac{d_{jj}d_{ki} -d_{ij}d_{kj}}{d_{ii}d_{jj} -d_{ij}d_{ij}}, \ \ \ \ \   \kappa_3 = \frac{d_{ii}d_{kj} -d_{ij}d_{ki}}{d_{ii}d_{jj} -d_{ij}d_{ij}}, \ \ \ \ \ 
\kappa_1 = 1 - \kappa_2 - \kappa_3. \\
\end{gathered}
\end{equation}
The barycenteric coefficients $\kappa_1$, $\kappa_2$, and $\kappa_3$ are in the range of $[0, 1]$ if the grid point $(x, y)$ is in the triangle.

\subsection{Wrapping Function}
\label{wrapping}
The wrapping function $\Psi: \bm{A}\in \mathbb{R}^{256 \times 256 \times 3} \rightarrow \bm{R} \in \mathbb{R}^{K \times 3}$ is defined as $\bm{R}(i) = \bm{A}(\bm{U}(i, 1), \bm{U}(i, 2))$, where $i$ is the index for the vertices of a 3D face. $\bm{R}(i)$ and $\bm{A}(\bm{U}(i, 1), \bm{U}(i, 2))$ are 3-dimensional vectors. $\bm{U} \in \mathbb{R}^{K\times 2}$ is the coordinates of shape in UV space from 3DMM \cite{blanz1999morphable}
. Again, bilinear interpolation is used if $\bm{U}(i, 1)$ or $\bm{U}(i, 2)$ is not an integer.


\newpage
\section{Experiments}
\subsection{Network Architecture}
\label{netarch}
We use standard encoder networks
for viewpoint, shape and illumination predictions, and a network similar to U-Net~\cite{ronneberger2015u} for reflectance prediction. The detailed configurations are given in Table~\ref{view_illum_shape}. Parameter $d$ is 7 for viewpoint network $f_{v}$ and 9 for illumination network $f_{\ell}$. Conv $3_{/2,1}$ denotes convoluitonal layer with kernel size of 3, where the stride and padding are 2 and 1, respectively. Each convolutional layer is followed by a Batch Normalization (BN) \cite{ioffe2015batch} layer and Rectified Linear Units (ReLU). Bilinear interpolation is adopted for the upsampling operation. Specifically, in Table \ref{view_illum_shape}, the layers in brackets are residual blocks. In Table \ref{reflectance}, we use shortcut to connect the feature maps of encoder and decoder, but different from U-Net, we use addition rather than concatenation to integrate information in the feature maps. For those encoder output shapes in brackets (\eg, ``[$128 \times 128 \times 64$]''), the feature map will be added as a shortcut to the decoder feature map (also with the same brackets).

\begin{table}[ht]
	\centering
	\footnotesize
	\setlength{\abovecaptionskip}{3pt}
    \setlength{\belowcaptionskip}{-5pt}
	\newcommand{\tabincell}[2]{\begin{tabular}{@{}#1@{}}#2\end{tabular}}
	\begin{tabular}{c|c|c|c|c|c}
		\hline 
		\multicolumn{3}{c|}{Viewpoint \& Illumination Network} & \multicolumn{3}{c}{Shape Network} \\ \hline
		Layer & Act. & Output shape & Layer & Act. & Output shape \\ \hline
		Input & - & $256 \times 256\times 3$ & Input & - & $256 \times 256\times 3$ \\
		Conv $4\times 4_{/2,1}$ & BN + ReLU & $128 \times 128 \times 32$ & Conv $4\times 4_{/2,1}$ & BN + ReLU & $128 \times 128 \times 64$ \\
		Conv $4\times 4_{/2,1}$ & BN + ReLU & $64 \times 64 \times 32$ & Conv $4\times 4_{/2,1}$ & BN + ReLU & $64 \times 64 \times 64$ \\
    	Conv $4\times 4_{/2,1}$ & BN + ReLU & $32 \times 32\times 64$ & Conv $4\times 4_{/2,1}$ & BN + ReLU & $32 \times 32\times 128$ \\ 
    	Conv $4\times 4_{/2,1}$ & BN + ReLU & $16 \times 16\times 64$ & Conv $4\times 4_{/2,1}$ & BN + ReLU & $16 \times 16\times 128$ \\
        \multirow{3}{*}{$\left[\begin{aligned} & \text{Conv} 3\times3_{/1,1} \\ & \text{Conv} 3\times3_{/1,1}\end{aligned}\right]$} & \multirow{3}{*}{$\begin{aligned} & \text{BN + ReLU} \\ & \text{BN + ReLU} \end{aligned}$} & \multirow{3}{*}{$\begin{aligned} & 16\times 16\times 64 \\ & 16\times 16\times 64 \end{aligned}$} & \multirow{3}{*}{$\left[\begin{aligned} & \text{Conv} 3\times3_{/1,1} \\ & \text{Conv} 3\times3_{/1,1}\end{aligned}\right]$} & \multirow{3}{*}{$\begin{aligned} & \text{BN + ReLU} \\ & \text{BN + ReLU} \end{aligned}$} & \multirow{3}{*}{$\begin{aligned} & 16\times 16\times 128 \\ & 16\times 16\times 128 \end{aligned}$} \\
		  &  &  &  &  & \\
		  &  &  &  &  & \\
		Conv $4\times 4_{/2,1}$ & BN + ReLU & $8\times 8\times 128$ & Conv $4\times 4_{/2,1}$ & BN + ReLU & $8\times 8\times 256$ \\ 
		\multirow{3}{*}{$\left[\begin{aligned} & \text{Conv} 3\times3_{/1,1} \\ & \text{Conv} 3\times3_{/1,1}\end{aligned}\right]$} & \multirow{3}{*}{$\begin{aligned} & \text{BN + ReLU} \\ & \text{BN + ReLU} \end{aligned}$} & \multirow{3}{*}{$\begin{aligned} & 8\times 8\times 128 \\ & 8\times 8\times 128 \end{aligned}$} & \multirow{3}{*}{$\left[\begin{aligned} & \text{Conv} 3\times3_{/1,1} \\ & \text{Conv} 3\times3_{/1,1}\end{aligned}\right]$} & \multirow{3}{*}{$\begin{aligned} & \text{BN + ReLU} \\ & \text{BN + ReLU} \end{aligned}$} & \multirow{3}{*}{$\begin{aligned} & 8\times 8\times 256 \\ & 8\times 8\times 256 \end{aligned}$}\\
		  &  &  &  &  & \\ 
		  &  &  &  &  & \\ 
		Conv $4\times 4_{/2,1}$ & BN + ReLU & $4\times 4\times 128$ & Conv $4\times 4_{/2,1}$ & BN + ReLU & $4\times 4\times 256$ \\
	    Conv $4\times 4_{/2,1}$ & - & $1\times 1\times d$ & Conv $4\times 4_{/2,1}$ & - & $1\times 1\times 228$ \\
		\hline
	\end{tabular}
	\renewcommand{\captionlabelfont}{\footnotesize}
	\caption{\footnotesize The detailed CNNs architectures of viewpoint, illumination, and shape networks.}\label{view_illum_shape}
\end{table}

\begin{table}[ht]
	\centering
	\footnotesize
	\setlength{\abovecaptionskip}{3pt}
    \setlength{\belowcaptionskip}{-5pt}
	\newcommand{\tabincell}[2]{\begin{tabular}{@{}#1@{}}#2\end{tabular}}
	\begin{tabular}{c|c|c|c|c|c}
		\hline 
		\multicolumn{6}{c}{Reflectance Network} \\
		\hline
		\multicolumn{3}{c|}{U-Net Encoder ($\downarrow$)} & \multicolumn{3}{c}{U-Net Decoder ($\uparrow$)} \\
		\hline
		Encoder Layer & Act. & Output shape & Decoder Layer & Act. & Output shape \\ \hline
		Input & - & $256 \times 256\times 3$ & Output & - & $256 \times 256\times 3$ \\
		- & - & - & Conv $3\times 3_{/1,1}$ & Tanh & $256 \times 256 \times 3$ \\
		- & - & - & Conv $3\times 3_{/1,1}$ & BN + ReLU & $256 \times 256 \times 3$ \\
		Conv $4\times 4_{/2,1}$ & BN + ReLU & $128 \times 128 \times 64$ & Upsample ($2\times$) & - & $256 \times 256 \times 64$ \\
		Conv $3\times 3_{/1,1}$ & BN + ReLU & [$128 \times 128 \times 64$] & Conv $3\times 3_{/1,1}$ & BN + ReLU & [$128 \times 128 \times 64$] \\
    	- & - & - & Conv $3\times 3_{/1,1}$ & BN + ReLU & $128 \times 128\times 64$ \\
    	Conv $4\times 4_{/2,1}$ & BN + ReLU & $64 \times 64\times 64$ & Upsample ($2\times$) & - & $128 \times 128 \times 64$ \\
    	Conv $3\times 3_{/1,1}$ & BN + ReLU & [$64 \times 64\times 64$] & Conv $3\times 3_{/1,1}$ & BN + ReLU & [$64 \times 64\times 64$] \\
    	- & - & - & Conv $3\times 3_{/1,1}$ & BN + ReLU & $64 \times 64\times 64$ \\ 
    	Conv $4\times 4_{/2,1}$ & BN + ReLU & $32 \times 32\times 128$ & Upsample ($2\times$) & - & $64 \times 64 \times 128$ \\ 
    	Conv $3\times 3_{/1,1}$ & BN + ReLU & [$32 \times 32\times 128$] & Conv $3\times 3_{/1,1}$ & BN + ReLU & [$32 \times 32\times 128$] \\
    	- & - & - & Conv $3\times 3_{/1,1}$ & BN + ReLU & $32 \times 32\times 128$ \\ 
    	Conv $4\times 4_{/2,1}$ & BN + ReLU & $16 \times 16\times 128$ & Upsample ($2\times$) & - & $32 \times 32 \times 128$ \\ 
    	Conv $3\times 3_{/1,1}$ & BN + ReLU & [$16 \times 16\times 128$] & Conv $3\times 3_{/1,1}$ & BN + ReLU & [$16 \times 16\times 128$] \\
    	- & - & - & Conv $3\times 3_{/1,1}$ & BN + ReLU & $16 \times 16 \times 128$ \\ 
    	Conv $4\times 4_{/2,1}$ & BN + ReLU & $8 \times 8\times 256$ & Upsample ($2\times$) & - & $16 \times 16 \times 256$ \\ 
    	Conv $3\times 3_{/1,1}$ & BN + ReLU & [$8 \times 8\times 256$] & Conv $3\times 3_{/1,1}$ & BN + ReLU & [$8 \times 8\times 256$] \\
    	Conv $4\times 4_{/2,1}$ & BN + ReLU & $4 \times 4\times 256$ & Conv $3\times 3_{/1,1}$ & BN + ReLU & $8 \times 8 \times 256$ \\ 
    	Conv $3\times 3_{/1,1}$ & BN + ReLU & $4 \times 4\times 256$ & Upsample ($2\times$) & - & $8 \times 8 \times 256$ \\
    	\hline
	\end{tabular}
	\renewcommand{\captionlabelfont}{\footnotesize}
	\caption{\footnotesize The detailed CNNs architectures of reflectance networks. Note that, the layers in the decoder (from input to output) are listed from bottom to top.}\label{reflectance}
\end{table}

\subsection{More Ablation Studies}
\label{moreablation}
\begin{figure}[t]
  \centering
  \renewcommand{\captionlabelfont}{\footnotesize}
  \includegraphics[width=6.5in]{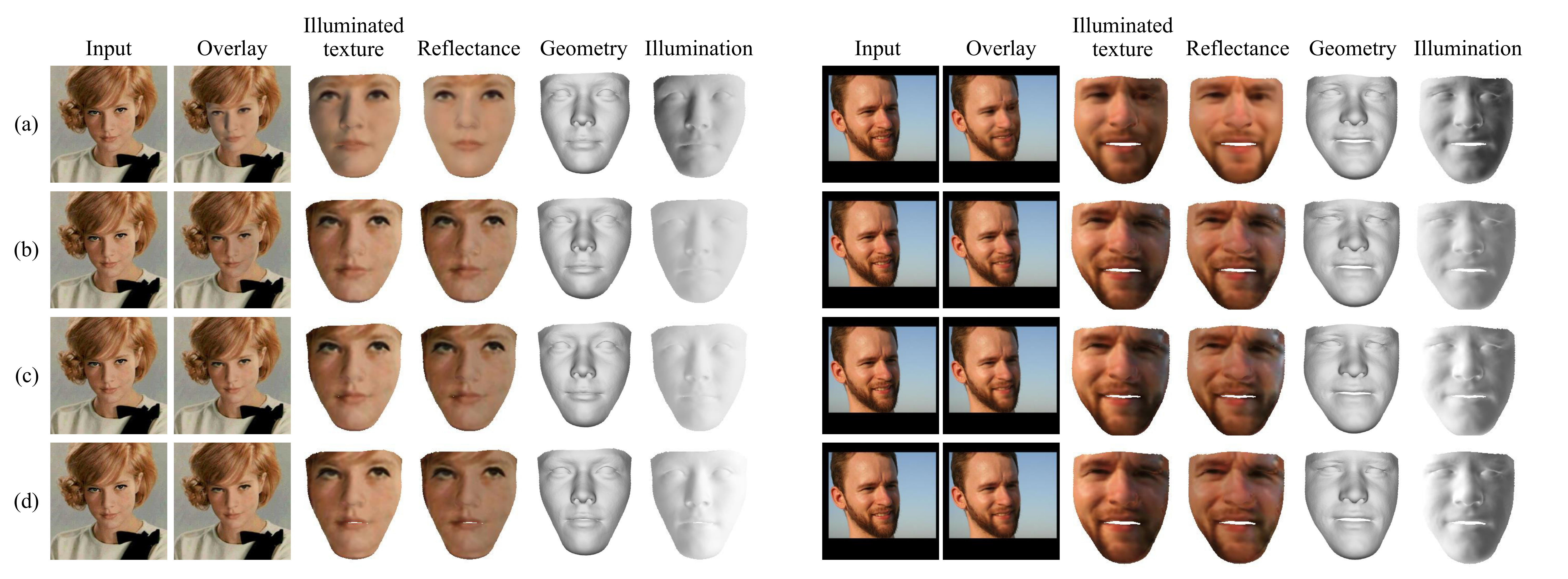}
  \caption{\footnotesize Ablations. (a) CEST with default settings. (b), (c) and (d) Averaged reflectance is used in training and the number of images from each video clips are 2, 4, and 8, respectively. {\color{gray}Errata: we mistakenly use the wrong image in the ICCV 2021 version. We have replaced it with the correct one in this version. Please refer to our latest arXiv version for up-to-date results.}}\label{fig:ablation2}
  \vspace{-0.1in}
\end{figure}

We perform more ablations for different settings of CEST. We explore the averaged representations, an approach adopted in \cite{tewari2019fml}, for reflectance consistency, where the averaged reflectance of a video clip is used to reconstruct the 3D face in each video frame. Here, we fix the size of minibatch, \ie 128, but vary the number of images from each video clip to 2, 4, and 8. Results are shown in Fig. \ref{fig:ablation2} (b), (c), and (d), respectively. As we can see, there are still some illumination in the reflectance, indicating that the averaged representation may not be a good strategy for learning disentangled facial parameters.

\begin{figure}[t]
  \centering
  \renewcommand{\captionlabelfont}{\footnotesize}
  \includegraphics[width=6.5in]{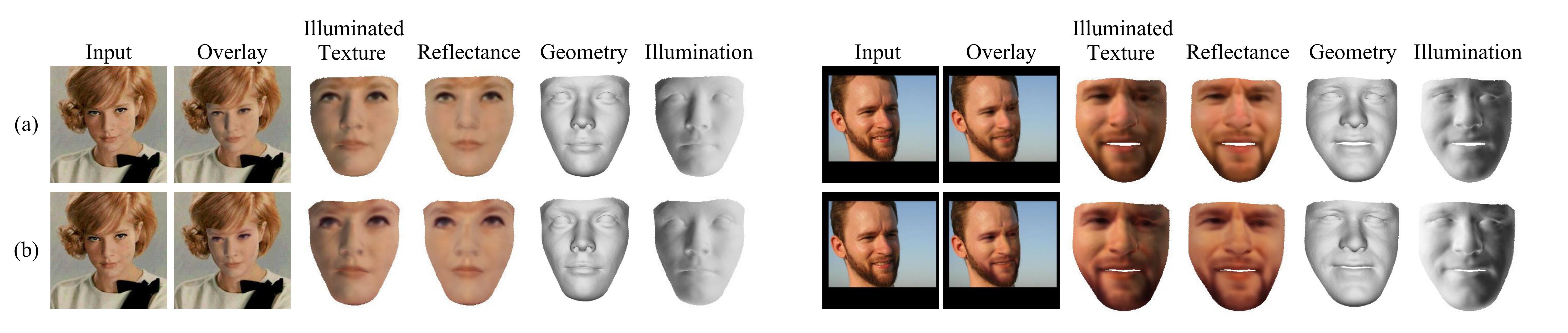}
  \caption{\footnotesize Ablations. (a) CEST with default settings. (b) Reflectance consistency is applied to videos, not video clips.}\label{fig:ablation3}
  \vspace{-0.1in}
\end{figure}
Fig. \ref{fig:ablation3} shows the results from CEST trained with reflectance consistency across video. The performance is comparable to those from CEST trained with default setting (reflectance consistency across video clip). It shows that consistency constraint can be generalized to longer videos if the recording environments are not changed dramatically.

\subsection{More Qualitative Comparisons}
\label{morequalitative}
In this section, we show more comparisons to the state-of-art methods \cite{booth20173d,sela2017unrestricted,richardson2017learning,thies2016face2face}. Since there is no publicly available implementations for these methods, we compare to the results presented in their papers.
\begin{figure}[t]
  \centering
  \renewcommand{\captionlabelfont}{\footnotesize}
  \includegraphics[width=6.5in]{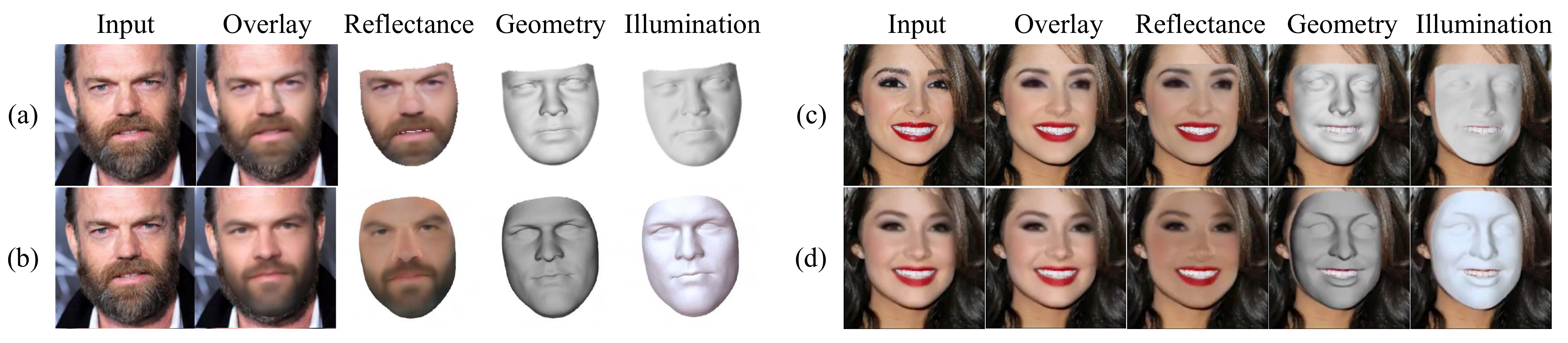}
  \caption{\footnotesize Comparisons to \cite{tewari2018self}. (a) and (c) Results from CEST. (b) and (d) Results from \cite{tewari2018self}.}\label{fig:250hz}
  \vspace{-0.1in}
\end{figure}

\begin{figure}[ht]
  \centering
   \setlength{\abovecaptionskip}{2pt}
   \setlength{\belowcaptionskip}{-5pt}
  \renewcommand{\captionlabelfont}{\footnotesize}
  \includegraphics[width=4.35in]{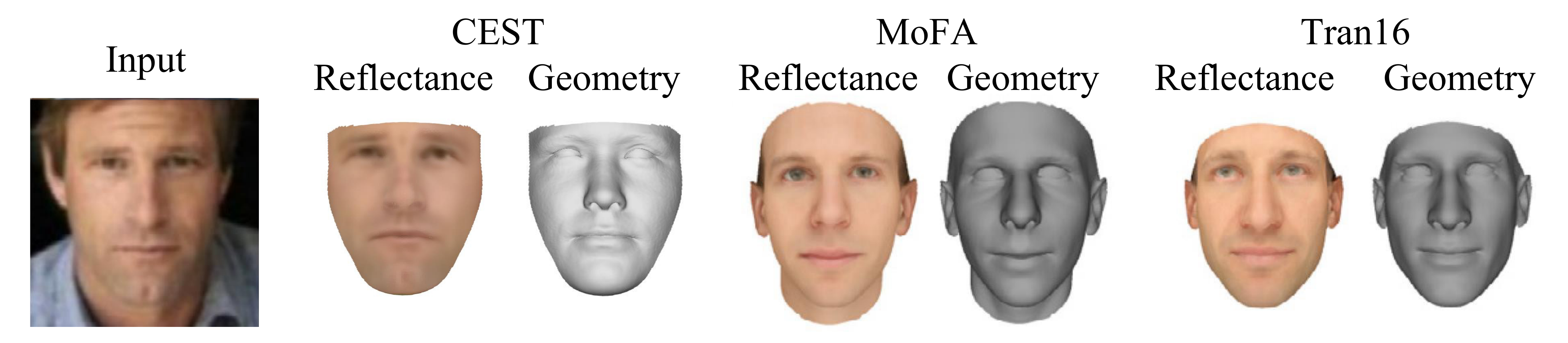}
  \caption{\footnotesize Comparisons to MoFA \cite{tewari2017mofa} and \cite{tuan2017regressing}.}\label{fig:mofa2}
  \vspace{-0.1in}
\end{figure}

\begin{figure}[ht]
  \centering
   \setlength{\abovecaptionskip}{2pt}
   \setlength{\belowcaptionskip}{-5pt}
  \renewcommand{\captionlabelfont}{\footnotesize}
  \includegraphics[width=4.3in]{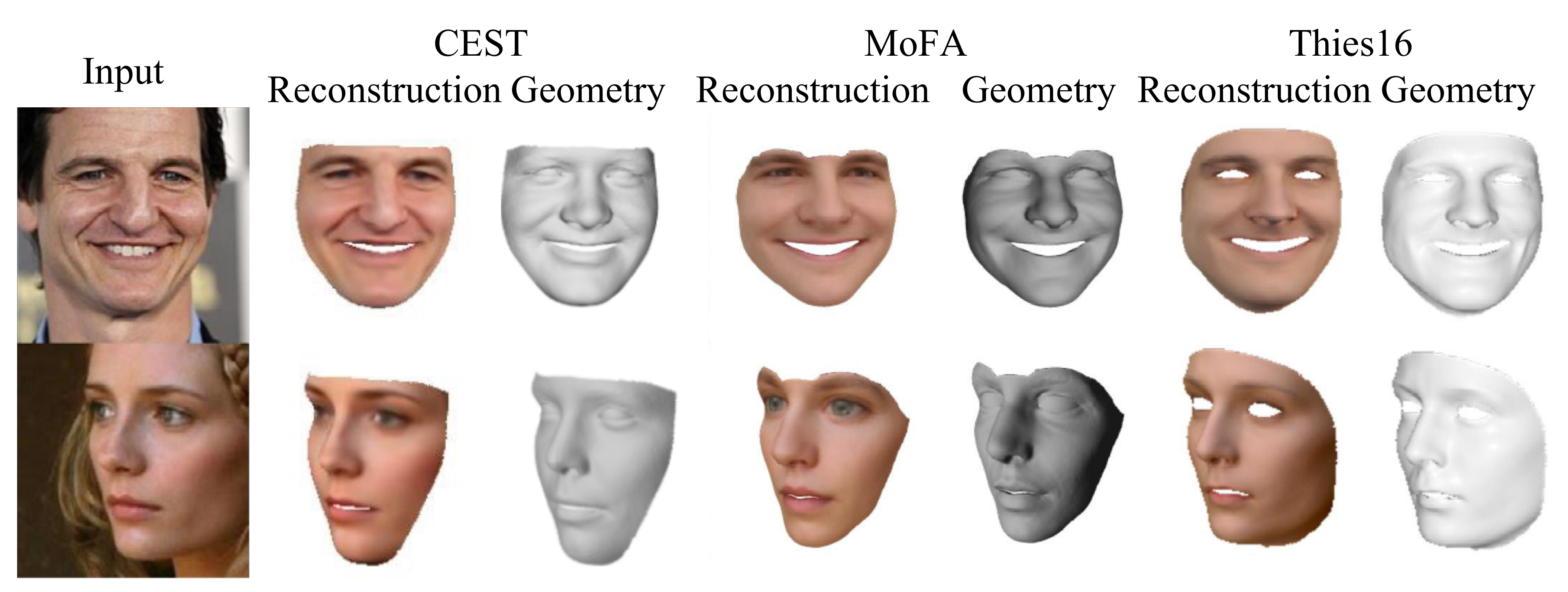}
  \caption{\footnotesize Comparisons to MoFA \cite{tewari2017mofa} and \cite{thies2016face2face}.}\label{fig:mofa3}
  \vspace{-0.1in}
\end{figure}

\begin{figure}[ht]
  \centering
   \setlength{\abovecaptionskip}{2pt}
   \setlength{\belowcaptionskip}{-5pt}
  \renewcommand{\captionlabelfont}{\footnotesize}
  \includegraphics[width=4.3in]{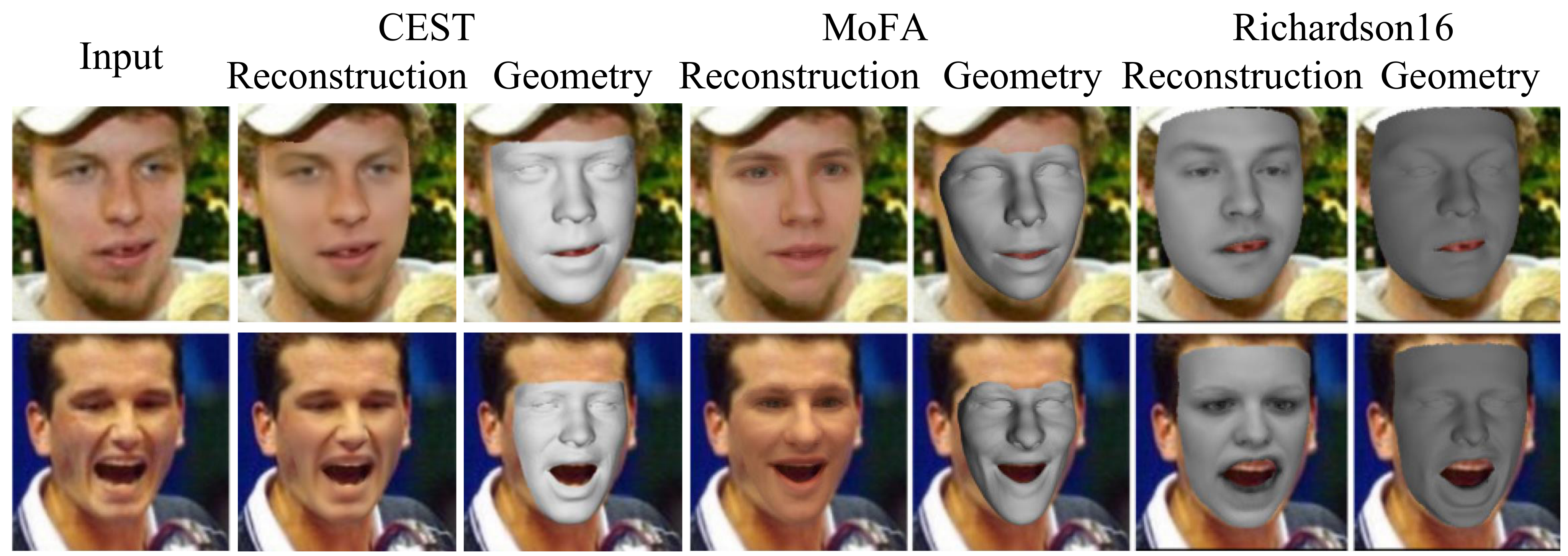}
  \caption{\footnotesize Comparisons to MoFA \cite{tewari2017mofa} and \cite{richardson2017learning}. Our estimated shapes show more accurate expressions.}\label{fig:mofa4}
  \vspace{-0.1in}
\end{figure}

\begin{figure}[ht]
  \centering
   \setlength{\abovecaptionskip}{2pt}
   \setlength{\belowcaptionskip}{-5pt}
  \renewcommand{\captionlabelfont}{\footnotesize}
  \includegraphics[width=4.3in]{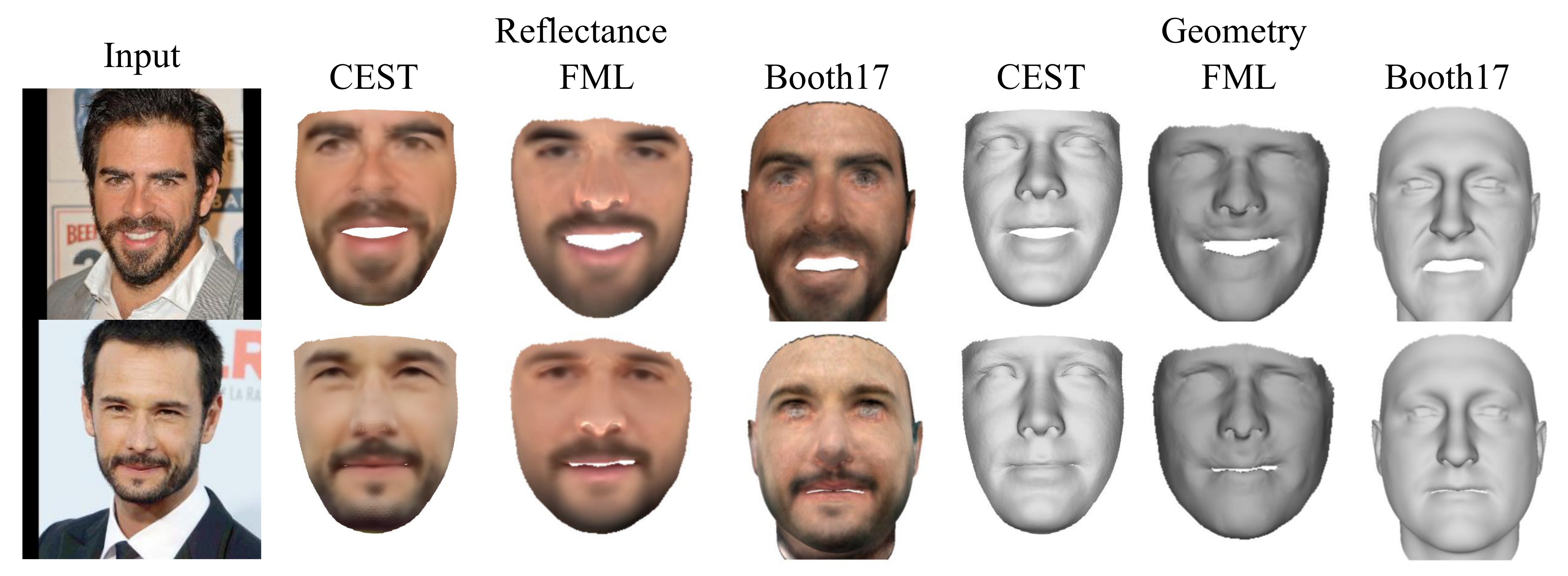}
  \caption{\footnotesize We compare CEST to FML \cite{tewari2019fml} and \cite{booth20173d}.}\label{fig:fml2}
\end{figure}

\begin{figure}[ht]
  \centering
  \renewcommand{\captionlabelfont}{\footnotesize}
  \includegraphics[width=5.5in]{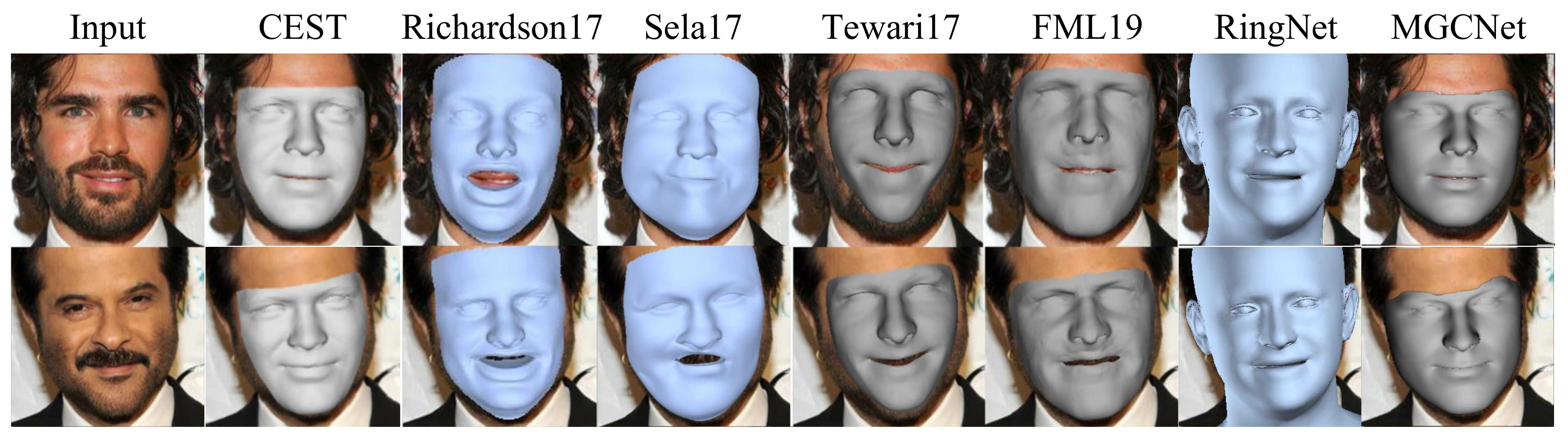}
  \caption{\footnotesize We compare the estimated shapes from CEST to those from \cite{richardson2017learning}, \cite{sela2017unrestricted}, \cite{tewari2017mofa}, \cite{tewari2019fml}, \cite{sanyal2019learning}, and \cite{shang2020self} (from left to right). Our estimated shapes are more stable and accurate.}\label{fig:fml3}
\end{figure}

Overall, CEST produces more stable and reasonable geometries, detailed reflectances, and realistic  reconstructions of the 3D faces. As shown in Fig. \ref{fig:250hz} (a) (b), Fig. \ref{fig:mofa4}, Fig. \ref{fig:fml2}, and Fig. \ref{fig:fml3}, the facial shapes predicted by CEST are more accurate in facial expressions and lip closure. In addition, the predicted reflectances show more personal characteristics, but less remaining illumination, as illustrated in Fig. \ref{fig:mofa2} and Fig. \ref{fig:fml2}. Lastly, CEST yields faithful 3D reconstructions, capturing more details than the other methods (see Fig \ref{fig:mofa3} and Fig \ref{fig:mofa4}).


\subsection{Challenging Cases}
\begin{figure}[ht]
  \centering
  \renewcommand{\captionlabelfont}{\footnotesize}
  \includegraphics[width=6.5in]{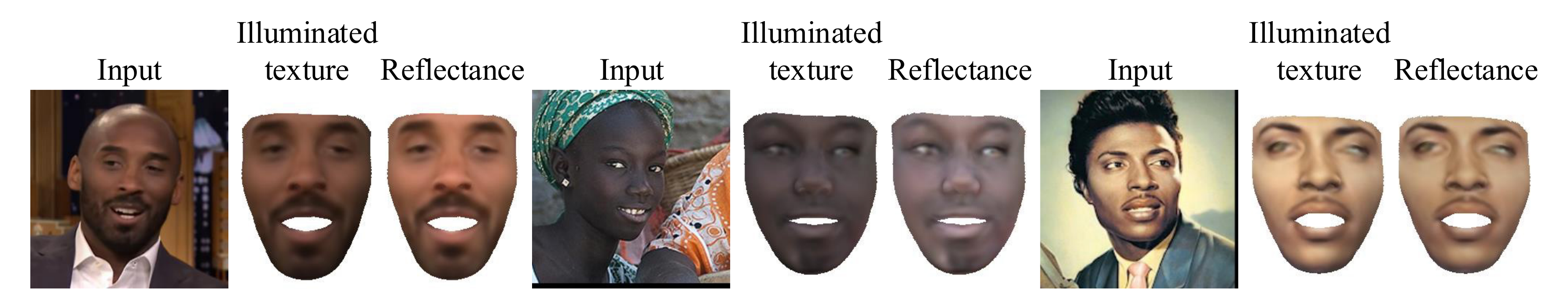}
  \caption{\footnotesize Some challenging examples.}\label{fig:hard_examples}
\end{figure}

We present some examples with dark skin in Fig. \ref{fig:hard_examples}. Although most people in the training set (VoxCeleb) are Caucasian, CEST still produces reasonable illumination and albedo for these examples. One limitation is that the reconstruction of the non-lambertian surface is inaccurate, e.g. eyes with unusual gaze directions.

\subsection{Photometric Error}
\label{morequantitative}

We compare CEST, IEST, FML~\cite{tewari2019fml} and Garrido~\cite{garrido2016corrective} on overlay face reconstruction. To measure the quality of the overlay images, we compute the average photometric error (R,G,B pixel values are from 0 to 255) between the input face image and the overlay face image. We experiment on 1,000 images in CelebA dataset~\cite{liu2015deep}. Table \ref{overlay} shows that the conditional estimation is beneficial for reconstructing the 3D face, and the proposed CEST outperforms existing methods by a large margin.

\begin{table}[h]
\renewcommand{\captionlabelfont}{\footnotesize}
\setlength{\abovecaptionskip}{2pt}
\footnotesize
\centering
\begin{tabular}{l|c|c|c|c}
\hline
Method & \textbf{CEST} & IEST & FML \cite{tewari2019fml} & Garrido16 \cite{garrido2016corrective} \\
\hline
Photometric Error & \textbf{10.74} & 13.76 & 20.65 & 21.95 \\
\hline
\end{tabular}
\caption{\footnotesize Photometric errors obtained by different methods.}\label{overlay}
\end{table}

\end{appendix}
\end{document}